\documentclass[acmsmall]{acmart}
\AtBeginDocument{%
	\providecommand\BibTeX{{%
			\normalfont B\kern-0.5em{\scshape i\kern-0.25em b}\kern-0.8em\TeX}}}

\setcopyright{acmcopyright}
\acmDOI{XXXXXXX.XXXXXXX}

\acmJournal{JACM}
\newenvironment{packed_itemize}{
\begin{list}{\labelitemi}{\leftmargin=1.5em}
  \setlength{\itemsep}{1pt}
  \setlength{\parskip}{0pt}
  \setlength{\parsep}{0pt}
  \setlength{\headsep}{0pt}
  \setlength{\topskip}{0pt}
  \setlength{\topmargin}{0pt}
  \setlength{\topsep}{0pt}
  \setlength{\partopsep}{0pt}
}{\end{list}}

\usepackage[normalem]{ulem} %to strike the words

\usepackage{soul}
\usepackage{url}
\usepackage{supertabular}
\usepackage{longtable} % Added By Chen Huiming
\usepackage{tabu}
\usepackage{textcomp}
\usepackage{makecell}

\newcommand{\para}[1]{{\vspace{4pt} \bf \noindent #1 \hspace{10pt}}}
\usepackage{color}
\usepackage{colortbl}

\renewcommand\arraystretch{1.1}
\newtheorem{mdefinition}{Definition}

\usepackage{amsmath,amsfonts,amsthm}
\usepackage{caption}
\usepackage{subcaption}
\usepackage{mathtools}
\usepackage{algorithm, algorithmic}
\usepackage{graphicx}
\usepackage{textcomp}
\usepackage{xcolor}
\usepackage{enumitem}
\usepackage{nicefrac}
\usepackage{stfloats}% <-- added
\usepackage[utf8]{inputenc}
\usepackage[english]{babel} 
\usepackage{multirow}
\usepackage{booktabs,tabularx} % Times Roman for text *and* math
\newcommand\norm[1]{\left\lVert#1\right\rVert}

\renewcommand{\underline}[1]{}

\newcommand{\tabincell}[2]{\begin{tabular}{@{}#1@{}}#2\end{tabular}}

\def\BibTeX{{\rm B\kern-.05em{\sc i\kern-.025em b}\kern-.08em
		T\kern-.1667em\lower.7ex\hbox{E}\kern-.125emX}}

\begin{document}
	
	\title{Advancements in Federated Learning: Models, Methods, and Privacy}
	
	%\email{webmaster@marysville-ohio.com}
	%\affiliation{%
		%	\institution{Institute for Clarity in Documentation}
		%	\streetaddress{P.O. Box 1212}
		%	\city{Dublin}
		%	\state{Ohio}
		%	\country{USA}
		%	\postcode{43017-6221}
		%}
	%
	%\author{Lars Th{\o}rv{\"a}ld}
	%\affiliation{%
		%	\institution{The Th{\o}rv{\"a}ld Group}
		%	\streetaddress{1 Th{\o}rv{\"a}ld Circle}
		%	\city{Hekla}
		%	\country{Iceland}}
	%\email{larst@affiliation.org}
	%\affiliation{%
		%	\institution{Inria Paris-Rocquencourt}
		%	\city{Rocquencourt}
		%	\country{France}
	%\affiliation{%
		%	\institution{Rajiv Gandhi University}
		%	\streetaddress{Rono-Hills}
		%	\city{Doimukh}
		%	\state{Arunachal Pradesh}
		%	\country{India}}
	%\affiliation{%
		%	\institution{Tsinghua University}
		%	\streetaddress{30 Shuangqing Rd}
		%	\city{Haidian Qu}
		%	\state{Beijing Shi}
		%	\country{China}}
	%\affiliation{%
		%	\institution{Palmer Research Laboratories}
		%	\streetaddress{8600 Datapoint Drive}
		%	\city{San Antonio}
		%	\state{Texas}
		%	\country{USA}
		%	\postcode{78229}}
	%\affiliation{%
		%	\institution{The Th{\o}rv{\"a}ld Group}
		%	\streetaddress{1 Th{\o}rv{\"a}ld Circle}
		%	\city{Hekla}
		%	\country{Iceland}}
	%\email{jsmith@affiliation.org}
	%\affiliation{%
		%	\institution{The Kumquat Consortium}
		%	\city{New York}
		%	\country{USA}}
	%\email{jpkumquat@consortium.net}
	\author{Huiming Chen}
	\email{chenhuiming@mail.tsinghua.edu.cn}
	\affiliation{%
		\institution{Department of Electronic Engineering, Tsinghua University}
		\city{Beijing}
		\country{China}
		\postcode{100084}
	}
\authornote{The first two authors have equal contributions.}

	\author{Huandong Wang}
	\affiliation{%
		\institution{Department of Electronic Engineering, Tsinghua University}
		\city{Beijing}
		\country{China}}
	\email{wanghuandong@tsinghua.edu.cn}
	\authornotemark[1]
	
	\author{Qingyue Long}
	\affiliation{%
		\institution{Department of Electronic Engineering, Tsinghua University}
		\city{Beijing}
		\country{China}}
	\email{longqy21@mails.tsinghua.edu.cn}

	\author{Depeng Jin}
	\affiliation{%
		\institution{Department of Electronic Engineering, Tsinghua University}
		\city{Beijing}
		\country{China}}
	\email{jindp@tsinghua.edu.cn}

		\author{Yong Li}
	\authornote{Correspondence author.}
	\affiliation{%
		\institution{Department of Electronic Engineering, Tsinghua University}
		\city{Beijing}
		\country{China}}
	\email{liyong07@tsinghua.edu.cn}

	%\affiliation{%
		%  \institution{Tsinghua University}
		%  \streetaddress{30 Shuangqing Rd}

	\renewcommand{\shortauthors}{H. Chen, et al.}
	
	\begin{abstract}
	 Federated learning (FL) is a promising technique for resolving the rising privacy and security concerns. Its main ingredient is to cooperatively learn the model among the distributed clients without uploading any sensitive data.  In this paper, we conducted a thorough review of the related works, following the development context and deeply mining the key technologies behind FL from the perspectives of theory and application. Specifically, we first classify the existing works in FL architecture based on the network topology of FL systems with detailed analysis and summarization.  Next, we abstract the current application problems, summarize the general techniques and frame the application problems into the general paradigm of FL base models. Moreover, we provide our proposed solutions for model training via FL. We have summarized and analyzed the existing FedOpt algorithms, and deeply revealed the algorithmic development principles of many first-order algorithms in depth, proposing a more generalized algorithm design framework. With the instantiation of these frameworks, FedOpt algorithms can be simply developed.	 As privacy and security is the fundamental requirement in FL, we provide the existing attack scenarios and the defense methods.  To the best of our knowledge, we are among the first tier to review the theoretical methodology and propose our strategies since there are very few works surveying the theoretical approaches. Our survey targets motivating the development of high-performance, privacy-preserving, and secure methods to integrate FL into real-world applications.
	\end{abstract}

	\begin{CCSXML}
<ccs2012>
   <concept>
       <concept_id>10003752.10010070</concept_id>
       <concept_desc>Theory of computation~Theory and algorithms for application domains</concept_desc>
       <concept_significance>500</concept_significance>
       </concept>
   <concept>
       <concept_id>10003033.10003034</concept_id>
       <concept_desc>Networks~Network architectures</concept_desc>
       <concept_significance>500</concept_significance>
       </concept>
   <concept>
       <concept_id>10010147.10010919</concept_id>
       <concept_desc>Computing methodologies~Distributed computing methodologies</concept_desc>
       <concept_significance>500</concept_significance>
       </concept>
   <concept>
       <concept_id>10002978</concept_id>
       <concept_desc>Security and privacy</concept_desc>
       <concept_significance>500</concept_significance>
       </concept>
 </ccs2012>
\end{CCSXML}

\ccsdesc[500]{Theory of computation~Theory and algorithms for application domains}
\ccsdesc[500]{Networks~Network architectures}
\ccsdesc[500]{Computing methodologies~Distributed computing methodologies}
\ccsdesc[500]{Security and privacy}
	
	\keywords{federated learning, architecture, communication efficiency, base models, distributed optimization, privacy and security }

	\maketitle
	\section{Introduction}
%\subsection{Background}
With the rapid development of information technology, a growing amount of data is generated by people's daily activities, which has become an invaluable asset in modern society.
Big data has created enormous benefits for society by providing businesses and individuals with new insights and foresights, but this reliance on data has also raised concerns about privacy and data security.
%While data has created huge benefits to society, contemporary world has been paying substantial attentions to the potential issues of privacy leakage and security breach (e.g., malicious attack in power grid  resulting in catastrophic blackout shown in Fig. \ref{fig:blackout}). 

In recent years, with the European Union's enactment of the General Data Protection Regulation \cite{GDPR} and legislative efforts in data security by the United States \cite{usprivacy} and China \cite{chinaprivacy}, data security and privacy protection have entered an important stage of development, i.e., the secure computation techniques including differential privacy, secure multi-party computing and FL  have been increasingly emphasized and widely applied in various industries.  

Federated learning provides an effective solution for ensuring compliant and secure data flow. As a new paradigm of machine learning, it is essentially a distributed machine learning framework that collaboratively solves a learning task by the federation of participating clients which are coordinated by a central server. Typically, the procedures of FL  involve client selection, broadcasting, the local computation with client's own data and global aggregation on the server \cite{ pmlr-v54-mcmahan17a}.   
It has been generally categorized into the following architectures \cite{10.1145/3298981}:
\begin{packed_itemize}
	\item
	Horizontal FL (HFL): HFL is the most common scenario in FL, where the data owners share the same feature space but different samples. HFL has been widely investigated since this scenario is widely popular in real applications. For example, with increasing health consciousness, a large amount of people wear smartwatch to monitor their health, these devices will generate massive data sharing the same features for  collaboratively learning the model together with coordination of the central server. 
	\item
	Vertical FL (VFL): VFL   is mainly dealing with the case that  the datasets over clients have different  feature spaces but share the same sample ID space.  It was first considered for two-party collaboration and recently has been extended to multiple-party scenario. VFL matches the real world applications very well. For example,  a patient can have the medical record in different hospitals and each may have distinct features. With many samples,  these hospitals will have more complete features and is able to jointly learn the model in a more accurate fashion. 
	\item
	Federated transfer learning (FTL):  FTL deals with the case that the dataset have both different feature spaces  and sample ID spaces. For example, in social computing network \cite{ZHANG202215}, it requires highly expensive and time consuming data labeling process. FTL techniques can be applied to alleviate the computation burden and provide the	solutions for the entire sample and feature space under a federation.
\end{packed_itemize}

Additionally, FL can conveniently merge multiple techniques for application scenario expansions, further promoting the industrial development. For example, FL has been applied in edge computing, which efficiently leverages the edge device's processing power to train the model, preventing the leakage of sensitive data without data transmission \cite{XIA2021100008}. Moreover, reinforcement learning has been incorporated for optimizing the resource allocations over both clients and servers \cite{9478223}. It has reached the incentive mechanism to encourage the clients' participation.

Currently, the field of FL has witnessed many outstanding research achievements, and these include performance \cite{ CFL_Sattler2021, SCAFFOLD, pmlr-v119-li20g} and security enhancements\cite{ sun2021defending, dubey2020differentially, geiping2020inverting}, etc. In terms of performance, FL frequently faces the issues of communication bottlenecks caused by  heterogeneous networks, physical distance, and communication overhead. For mitigating these issues, novel FL algorithms have been investigated for improving the efficiency of training complex models. They include local and global acceleration strategies, asynchronous coordination, compression and sparsification. Regarding security issues, since FL requires participation from multiple parties, it may introduce extra security risks. In recent years, new solutions and technologies have emerged for enhancing security defense capabilities, resisting data leakage and backdoor attacks that may occur during the FL process. In summary, It is foreseeable that the huge demand for secure computing will drive the continuous and fast development of FL, including:
\begin{packed_itemize}
	\item
\textbf{The continuous improvement of the system and increasing unification of industry standards:} the application of FL requires standardized specifications. With the rapidly increasing implementation, standard formulation work is also in progress. Various international organizations have started developing the standards of FL since 2018. Currently, IEEE and ITU-T have released framework and function standards \cite{iu-t,9382202}. Moreover, the security requirement standards and technical application standards for FL with multi-party secure computation  are also in the process of being drafted.
	\item
\textbf{Increasingly diverse application scenarios and growing number of applications:} many traditional industries that urgently require digital transformation have the characteristics of data-intensive, such as the financial industry and internet of things. They generally have large-scale application scenarios for data flow and put forward more stringent requirements for security management. These traditional industries are in an urgent need to establish the secure and private data flow to achieve effective data fusion and utilization. Therefore, FL can be a fit for ensuring the secure and compliant data flow and promotes the digital development of traditional industries.
\end{packed_itemize}   
On the one hand, both the developments of applications and techniques in FL become increasingly mature, and on the other hand, the problems faced in FL applications are also becoming increasingly complicated. Therefore, our survey targets at answering the following questions:
\begin{packed_itemize}
	\item
	How to abstract problems from a specified actual application and propose solutions for FL?
	\item
	How to deploy multiple (semi-) decentralized parameter servers to establish a (semi-) decentralized FL system?
	\item How to build an efficient, secure, trustworthy, and scalable FL system?
\end{packed_itemize}

\scriptsize
\begin{table*}[ht]
	%  \resizebox{1\textwidth}{!}{%
		\begin{center}
			\begin{tabular}{l|l}
				\toprule
				\textbf{Paper} & \textbf{Description} \\\hline
				%\midrule
				\cite{10.1145/3298981}        &    \tabincell{l}{Concepts and applications of FL in terms of its definition, privacy techniques, and\\ categories, including horizontal FL, vertical FL, and federated transfer learning.}  \\ \hline
				\cite{niknam2020federated}      &   \tabincell{l}{Federated machine learning in wireless networks in terms of its applications and challenges.}   \\ \hline
				\cite{li2020federated}          &    \tabincell{l}{Unique characteristics and challenges of FL, including expensive communication, systems heterogeneity, \\statistical heterogeneity, and privacy concerns, and existing studies and directions to solve these challenges. }  \\ \hline
				\cite{Wei2020Federated}         &  \tabincell{l}{Survey on FL implementing in mobile edge networks in terms of several\\ challenges, including communication costs, resource allocation, and privacy and security.}\\ \hline
				\cite{kairouz2019advances}      & Survey on comprehensive FL.\\ \hline
				%\bottomrule
			\end{tabular}
		\end{center}
		%}
	\caption{Selected surveys of FL.}
	\label{Tab:existingsurvey}
\end{table*}
\normalsize

%\subsection{Related Surveys}

\para{Differences between existing surveys.}
There have been a number of existing surveys about FL techniques. Next, we provide the introduction of the representative surveys. As the pioneer FL survey,
Yang et al.~\cite{10.1145/3298981} introduce the concepts and applications of FL in terms of its definition, privacy techniques, and categories, including horizontal FL, vertical FL, and federated transfer learning.
Niknam et al.~\cite{niknam2020federated} investigate federated machine learning in wireless networks in terms of its applications and challenges.
Li et al.~\cite{li2020federated} focus on unique characteristics and challenges of FL, including expensive communication, systems heterogeneity, statistical heterogeneity, and privacy concerns, and existing studies and directions to solve these challenges. 
Wei et al.~\cite{Wei2020Federated} focus on FL implementing in mobile edge networks in terms of several challenges, including communication costs, resource allocation, and privacy and security.
%Different with them, we focus on optimization algorithms.
Kairouz et al.~\cite{kairouz2019advances} focus on the privacy and security issues in the FL system, and they mainly summarize existing studies in terms of attack methods and defence methods.

Different from existing surveys, we have a thoughtful summary and discussion of existing literature, which not only digs deep into the existing perspectives, but also provides analysis from different perspectives.
Specifically, in terms of the FL architecture, based on the existing taxonomy, we further focus on the topological structures of the FL systems. Moreover, based on the seldom surveys on FedOpt \cite{DBLP:journals/corr/abs-2107-06917}, we have summarized and analyzed the existing FedOpt algorithms, which play a key role in improving performance. In particular for many first-order FedOpt, we  provide the general frameworks including local and global acceleration strategies. In addition, based on the existing surveys on FL applications, we have provided a generalized federated base models and offered our solutions to facilitate larger-scale applications.

\begin{figure*}
	\centering
	\includegraphics[width=1\linewidth]{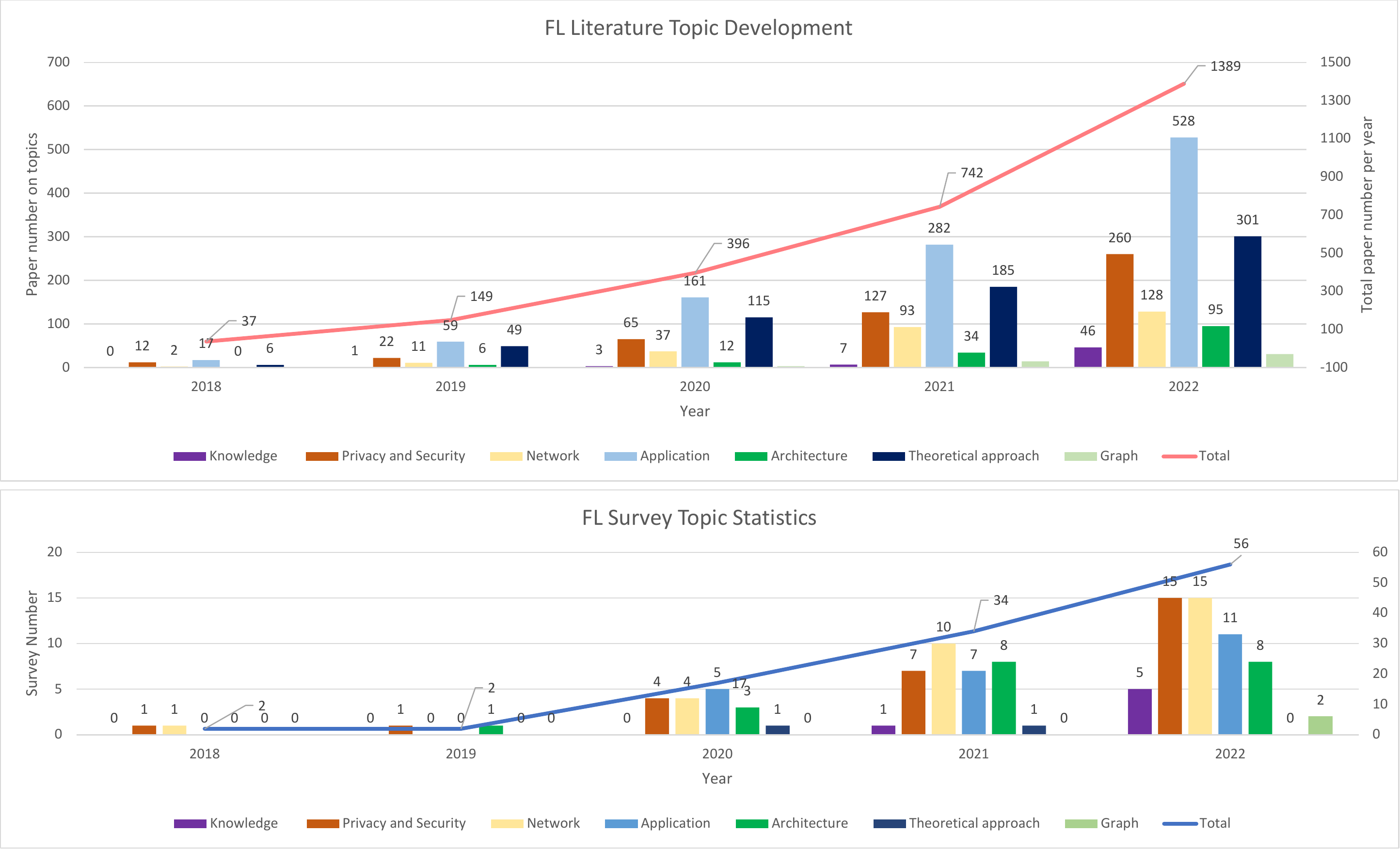}
	\caption{The statistics on FL related topic papers from years 2018 to 2022. It can be seen FL has been under fast development and can be expected to maintain quick growth in near future. Specifically,  the topics of privacy and security, theoretical approaches (optimization in majority) and applications dominates this field, the surveys on privacy and security and applications follows this pattern while this area significantly lacks the optimization related surveys.}
	\label{fig:surveystatistics}
\end{figure*}

\para{Our contributions.}
%\subsection{Our Contributionss}
%While most existing surveys provide prominent reviews from the systematic perspectives, the basic theories of FL, such as theoretical approaches which play a fundamental role in the FL development, have less been investigated in detail. 
In this survey, we provide a comprehensive and motivative review for targeting at developing high-performance and privacy-preserving approaches, as to integrate FL into real-world applications. To be specific, from theoretical perspectives, we review the architectures and optimization methods, both of which play a fundamental role in the performance. We review the FL applications, abstract the federated base models underlying them, and propose our solutions. Moreover, privacy and security are investigated with the existing attack scenarios and the defense methods. Finally, we classify the existing works in applications and project our insights into challenges and opportunities.
Overall, we summarize the contributions we make in this survey as follows:

\begin{packed_itemize}
    \item\textbf{New summarization on FL trends:} We have conducted the extensive review  in the topics of FL research papers and surveys from the year 2018 to 2022, {\textbf{including 2,713 research papers and 93 surveys}}, and the results are shown in Fig. \ref{fig:surveystatistics}. In general, we can draw the conclusions as follows:
    \begin{packed_itemize}
    	\item %\textbf{Major Topic Classification:}  
        \textbf{A novel taxonomy:} 
        With elaborately partitioning strategy to show the development of FL in each topic, we can classify the existing FL works into knowledge domain, privacy and security, network, real application, architecture, theoretical approach, and graph domain. In particular, optimization theory dominates the theoretical approach in FL.  
    	\item \textbf{Rapid development:} FL has been maintaining a fast development speed. In particular, there are 1,389 FL papers released in 2022, which is nearly the sum of  works in previous years.  Especially, with the development of smart city worldwide, the FL applications will maintain a fast growth speed, which will also drive the theoretical approach development.  
    	\item \textbf{Lack of surveys in theoretical approaches:} We also find the FL field lacks significantly the related surveys in theoretical approaches, while they are the underlying basis supporting the FL development. Although the year 2022 has witnessed FL surveys becoming booming on broad topics, there are only exactly 2 surveys on the theoretical approach topic.     
    \end{packed_itemize}
    
    \item \textbf{Federated base  model abstraction and our proposed solutions:} From Fig. \ref{fig:surveystatistics}, we can see FL applications constitute a significant proportion in the studies on FL. We project our insights into these applications and abstract the problems, then we have thoroughly analyzed and classified these problems with federated solutions. In general, the underlying issues of these applications can be reduced to the popular machine learning model training,  non-smooth regularization, multi-task learning, and matrix factorization, in the context of FL.  \textbf{It is worth emphasizing that we have provided novel  approaches for  learning these models under federated settings. }
    \item \textbf{Proposed algorithmic design frameworks:} FedOpt algorithms play the fundamental role in FL and they also occupy a significant proportion in the research efforts of FL. Therefore, we carry out a comprehensive review of the works in FedOpt algorithms, delving deeply into the generalization of the common  patterns underlying these algorithms. Specifically, the FedOpt algorithms are identified into first-order (majority in FedOpt), second-order and ADMM based scenarios. In particular, we further generalize the existing first-order works into the local acceleration and global acceleration frameworks respectively, which can be simply instantiated for algorithm development. \textbf{To the best of our knowledge, we are among the first-tier for comprehensively surveying the theoretical approaches in FedOpt.}

\end{packed_itemize}
The rest of this survey is organized as follows: Section \ref{arch} discusses the FL architectures. Section \ref{apa} illustrates the application problem abstraction. In Section \ref{opt},  FedOpt algorithms are reviewed and we also provide the instantiations in the supplementary material.  Section \ref{pas} studies the privacy and security issues. Section \ref{app} presents both the technical and real applications. Section \ref{cao} illustrate the challenges and future directions.

\begin{figure*}
	\centering
	\includegraphics[width=1\linewidth]{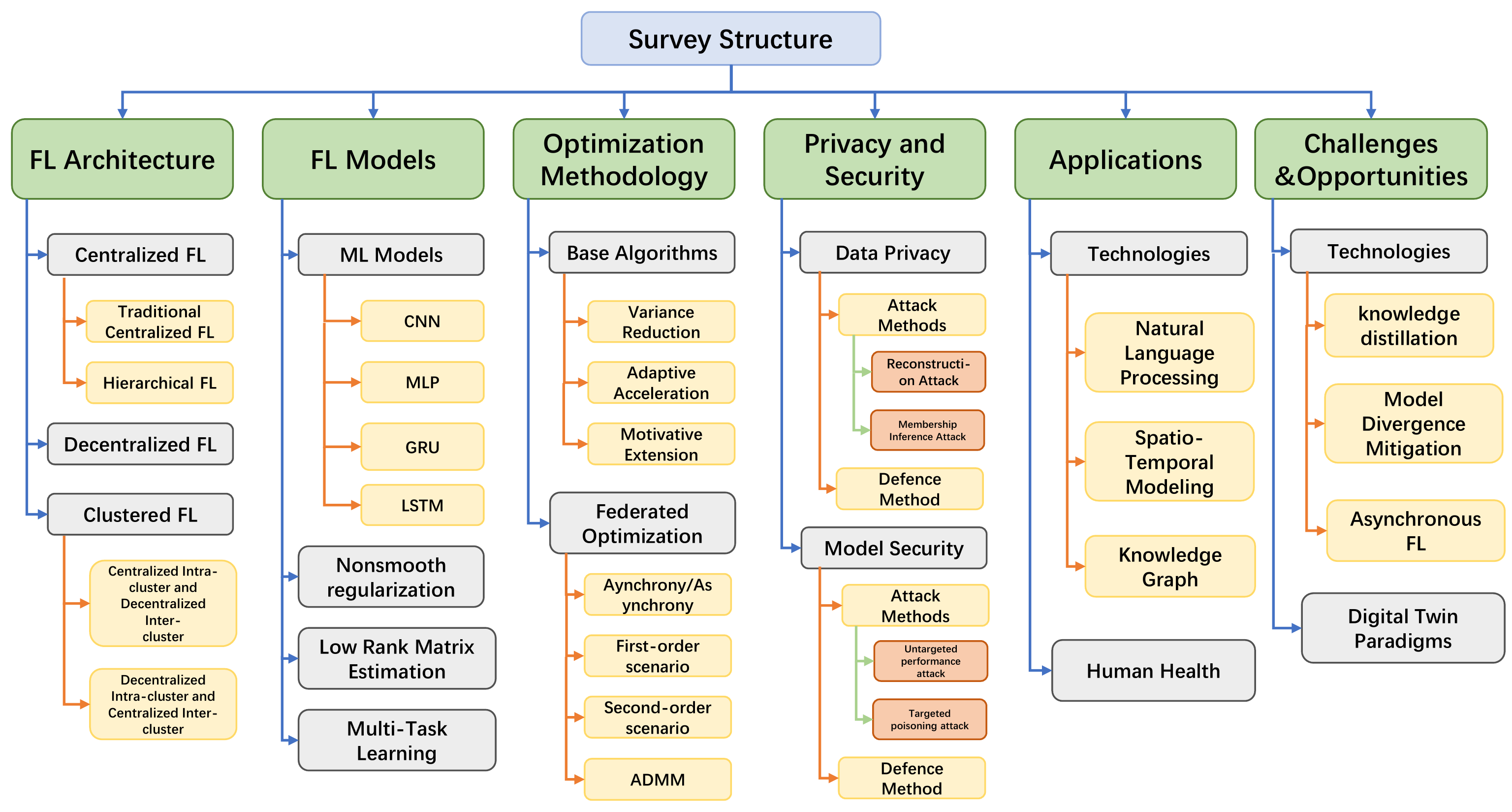}
	\caption{Organization of this survey.}
	\label{fig:sueveystructure}
\end{figure*}

%架构改一下，以centerilized/De-C为主，H/V为辅

	\section{Architecture}\label{arch}
FL architecture plays the fundamental role in FL \cite{survey_2021}, building FL architectures can be  effective and practical for improving the performance. Based on {the distribution characteristics of the data}, three types of FL techniques have been identified: horizontal FL, vertical FL, and federated transfer learning \cite{10.1145/3298981}. {Different from} this type of categorization, our main focus is on the topological architecture of the network of participating devices in FL.
While typical FL architecture consists of a central server and a number of clients, with the central server coordinating clients for cooperative learning \cite{li2020federated,survey_2021}, the decentralization \cite{DFL_Taya2022} or  hierarchy \cite{HFL_Qu2022} have  become increasingly popular recently. Therefore, we will categorize the FL architectures  into three scenarios: centralized FL, decentralized FL and clustered FL.
%We will first introduce the centralized topological structures of FL. The classical FL with one central server and numerous local clients together with the hierarchical FL belongs to the centralized FL.
%The centralized topological structures of FL will be discussed initially. Centralized FL includes hierarchical FL and the traditional FL, which has a single central server and multiple local clients. Then, we focus on the decentralized FL, which does not contain a central server. At last, the clustered FL is discussed as a hybrid of centralized and decentralized FL.

\subsection{Centralized FL}
Centralized structures are the most common topology for an FL system, where a centralized server is in charge of coordinating different clients. It was first introduced in \cite{pmlr-v54-mcmahan17a}, where each client trains the local model  with the its data and communicates with the server for  aggregation \cite{Yang_2021}. 

\para{Traditional FL.}The traditional centralized FL has been among the most popular choice in real applications. Its major bottleneck is the communication overhead. For mitigating this issue, Yang et al. \cite{OTA_TWC2020} combine device selection and beamforming design, and investigate the superposition property of a wireless MAC for achieving the fast global model aggregation. By considering the characteristics of the wireless channel, Xu et al. \cite{OTA_JSAC1} propose a modified FedAvg algorithm with a dynamic learning rate that is able accommodate the fading channel. As for the model aggregation problem, Fan et al. \cite{OTA_JSAC2} first characterize the intrinsic temporal structure of the model aggregation series via a Markovian probability model and develop a message passing based algorithm with low complexity and near-optimal performance. For further communication efficiency, Jing et al. \cite{OTA_TWC2022} adopt statistical channel state information. However, traditional centralized FL  still exists other challenges including statistical heterogeneity, computation overhead, security and privacy \cite{survey_2021}.

\para{Hierarchical FL.}This scenario generally exists in the well-known cloud-edge-client system and is organized in a tree structure \cite{HFL_Shi2021}, the major goal is to alleviate the above issues in traditional FL \cite{HFL_liuICC2020}.  Specifically,  the cloud layer coordinates  the edge layer, while the edge layer cooperates with the terminal layer. Since the edge layer has taken on the communication overhead from the cloud layer and the computation from the terminal layer \cite{10.1145/3514501}, this architecture has been shown the capability of high communication efficiency and low latency in the asynchronous and heterogeneous system \cite{AHFL_Wang2022, HFL_Zheng2021}. For further improving the system robustness, wireless channel has been considered for  hierarchical over-the-air FL (HOTAFL) \cite{HFL_aygun2022}. Liu et al. \cite{HFL_liuTWC2022} and Xu et al. \cite{HFLXu2022} investigate the issues of joint user association and wireless resource allocation in HOTAFL.

\subsection{Decentralized FL}

In practice, the centralized FL is vulnerable when suffering from server malfunctions, malicious attacks and untrustworthy servers. For addressing these issues, Li et al. \cite{DBLP:journals/tpds/LiSWDMSHP22} propose a decentralized framework with the integration of the blockchain to delegate the responsibility of storing the model to the nodes on the network, it also  optimizes the computing resource allocation.  Liu et al. \cite{DFL_Liu2022} suggest a generic decentralized FL framework , which implements both multiple local updates and multiple inter-node communications with the compression technique periodically.  Hashemi et al. \cite{DFL_Hashemi2022} propose DeLi-CoCo and demonstrate how the decentralized FL benefiting from additional gossip stages between gradient iterations for faster convergence. For further improving the performance, Li et al. \cite{DFL_Li2022} propose a Def-KT algorithm to fuse models among clients by transferring their learned knowledge to each other and has alleviated the client-drift issue. Xiao et al. \cite{DFL_Xiao2021} develop a  decentralized FL framework with an inexact stochastic parallel random walk, which can be partially immune to the time-varying dynamic
network.

In terms of security, since decentralized clients can only talk to their neighbors, the dissensus attack can poison the clients' collaboration.  To overcome this issue,  He et al. \cite{DBLP:journals/corr/abs-2202-01545} propose a Self-Centered Clipping (SCClip) algorithm for Byzantine-robust consensus and optimization. Che et al. \cite{DFL_Che2022} adopt the committee mechanism to track the uploaded local gradients, which has shown both performance and security improvement.  

\tiny
\begin{table*}[ht]
	%  \resizebox{1\textwidth}{!}{%
		\begin{center}
			\begin{tabular}{l|l|l|l}
				\toprule
				\textbf{Categories} & \textbf{Topological structure} & \textbf{Feature} &\textbf{Ref.}   \\
				\midrule
				\multirow{2}{*}{Centralized FL}
				& Traditional Centralized FL  & One centralized server aggregates all model parameters updated by clients. & \makecell[l]{\cite{pmlr-v54-mcmahan17a}, \cite{Yang_2021}, \cite{OTA_TWC2020},   \cite{OTA_JSAC1}, \\  \cite{OTA_JSAC2}}   \\
				\cline{2-4}
				& Hierarchical FL  & \makecell[l]{The client-edge-cloud structure: \\One cloud server is connected to edge servers in the middle layer.  \\ Every edge server connects to partial clients in the bottom layer.} & \makecell[l]{\cite{HFL_liuICC2020}, \cite{HFL_Shi2021}, \cite{AHFL_Wang2022},  \cite{HFL_Zheng2021}, \\ \cite{HFL_Abad2020},  \cite{HFL_aygun2022},   \cite{HFL_liuTWC2022}, \cite{HFLXu2022},\\ \cite{HFL_Rango2021},  \cite{HFL_Zhong2022}}.  \\
				\cline{1-4}
				\multirow{1}{*}{Decentralized FL}
				& Decentralized FL   &  \makecell[l]{Without central server.\\ All clients communicate with each other in a distributed way.} & \makecell[l]{ \cite{DFL_Kim2020}, \cite{DFL_Kor2020}, \cite{DFL_Tseng2022}, \cite{DFL_Liu2022}, \\ \cite{DFL_Hashemi2022},  \cite{DFL_GHolami2022},  \cite{DFL_Gouissem2022}, \cite{DFL_Che2022}, \\
					\cite{DFL_Li2022}, \cite{DFL_Xiao2021} }.   \\
				\cline{1-4}
				\multirow{4}{*}{Clustered FL}
				& \makecell[l]{ Centralized Intra-cluster, \\
					Decentralized Inter-cluster}    & \makecell[l]{Some client are connected with one edge server.\\ The edge servers are decentraliedly connected with each other.} & \cite{CFL_GouICC2022}, \cite{DFL_Sun2022}.   \\
				\cline{2-4}
				&\makecell[l]{Decentralized Intra-cluster,   \\ Centralized Inter-cluster}  & \makecell[l]{Each client is decentraliedly connected with each other.\\ Global aggregations is aggregated by a central server.}  & \cite{CFL_NIPS2020},\cite{DFL_Lin2021},  \cite{CFL_LiIOT2022}.  \\
				\bottomrule
			\end{tabular}
		\end{center}
		%}
	\caption{Existing architectures in FL.}
	\label{Tab:architecture}
\end{table*}
\normalsize

\subsection{ Clustered FL }
In this scenario, users are distributed and partitioned into clusters. It exploits geometric properties of the FL loss surface to group the client population into clusters with jointly trainable data distributions \cite{CFL_NIPS2020,CFL_Sattler2021}. In general, we can regard clustered FL as a hybrid of centralized and decentralized FL or semi-decentralized FL. 

%Due to the classic centralized FL system has a high dependence on the central server, causing unguaranteed reliability of the central server and extremely high consumption of communication resources \cite{CFL_GouICC2022}. Meanwhile, the decentralized FL system is vulnerable to attack from malicious clients and difficult to allocate resources uniformly. \cite{CFL_NIPS2020} firstly proposed the clustered FL, where users are distributed and partitioned into clusters. It exploits geometric properties of the FL loss surface to group the client population into clusters with jointly trainable data distributions \cite{CFL_Sattler2021}. Due to the fact that clustered FL has two primary architectures, clustered FL is regarded as a hybrid of centralized and decentralized FL or semi-decentralized FL.

\para{Centralized Intra-cluster and Decentralized Inter-cluster.}In this architecture, each client node is associated with an edge server according to the specific predefined criteria, and the edge servers are deployed in a decentralized manner. As an exemplar, when this architecture is applied in Internet of Vehicles (IoV), vehicles in each cluster  train their models and communicates with the service provider (SP) for the aggregation. Then, the SPs are deployed in decentralized manner and they communicate with their neighboring SPs for information exchange. Here, communication strategy plays the crucial role in the performance. Gou et al. \cite{CFL_GouICC2022} apply gossip protocol for communication between clusters and introduce the leader-follower strategy to make full use of bandwidth and reach a faster convergence. 

\begin{figure*} 
	\begin{minipage}[t]{0.48\linewidth} % 如果一行放2个图，用0.5，如果3个图，用0.33
		\centering
		\begin{subfigure}{1\textwidth}
			\centering
			\includegraphics[width=0.9\linewidth]{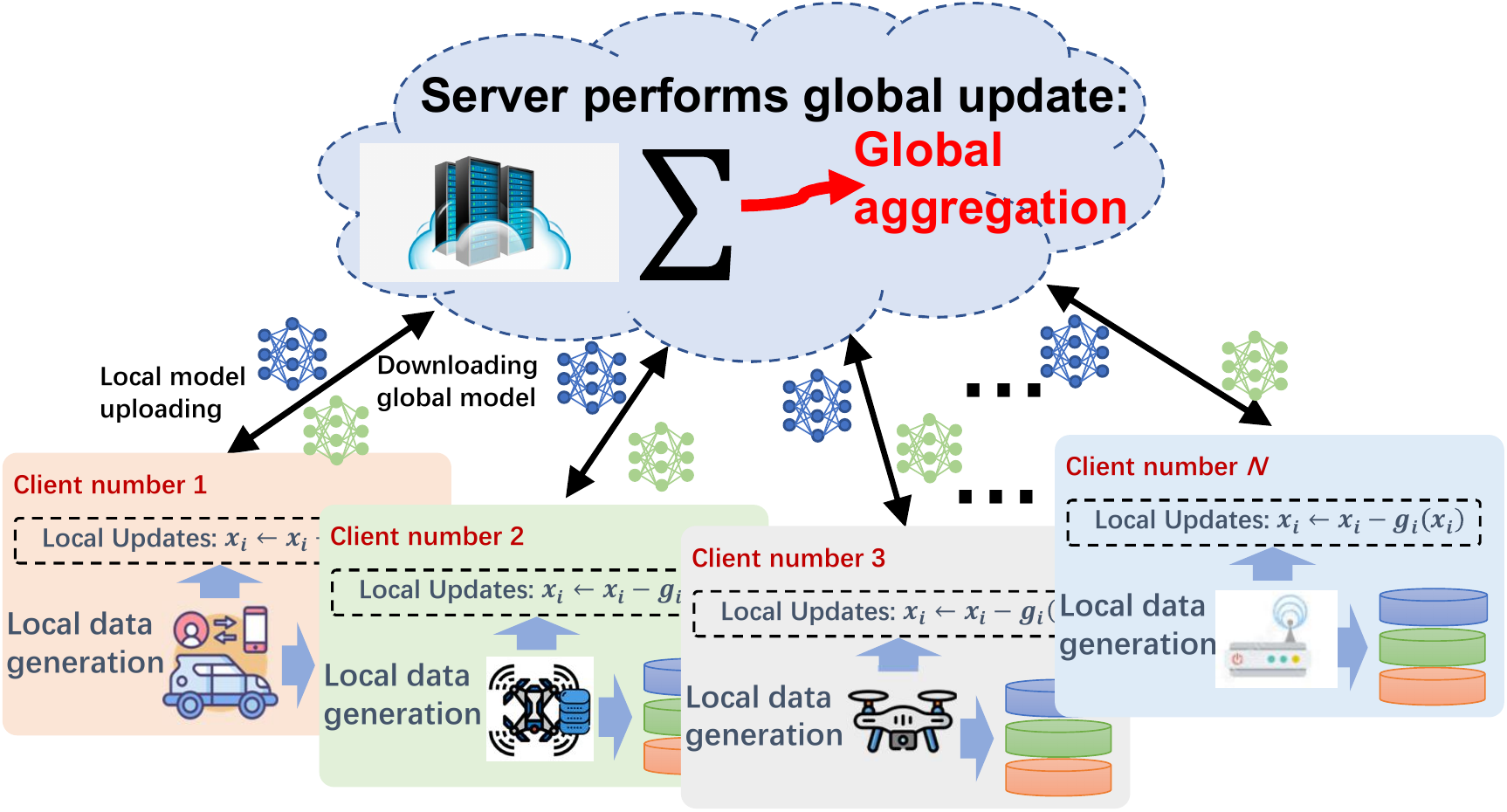}
		\end{subfigure}\\
	    \subcaption{Centralized FL architecture.}
	\end{minipage} \hfill
	\begin{minipage}[t]{0.48\linewidth} % 如果一行放2个图，用0.5，如果3个图，用0.33
		\centering
		\begin{subfigure}{1\textwidth}
			\centering
			\includegraphics[width=0.9\linewidth]{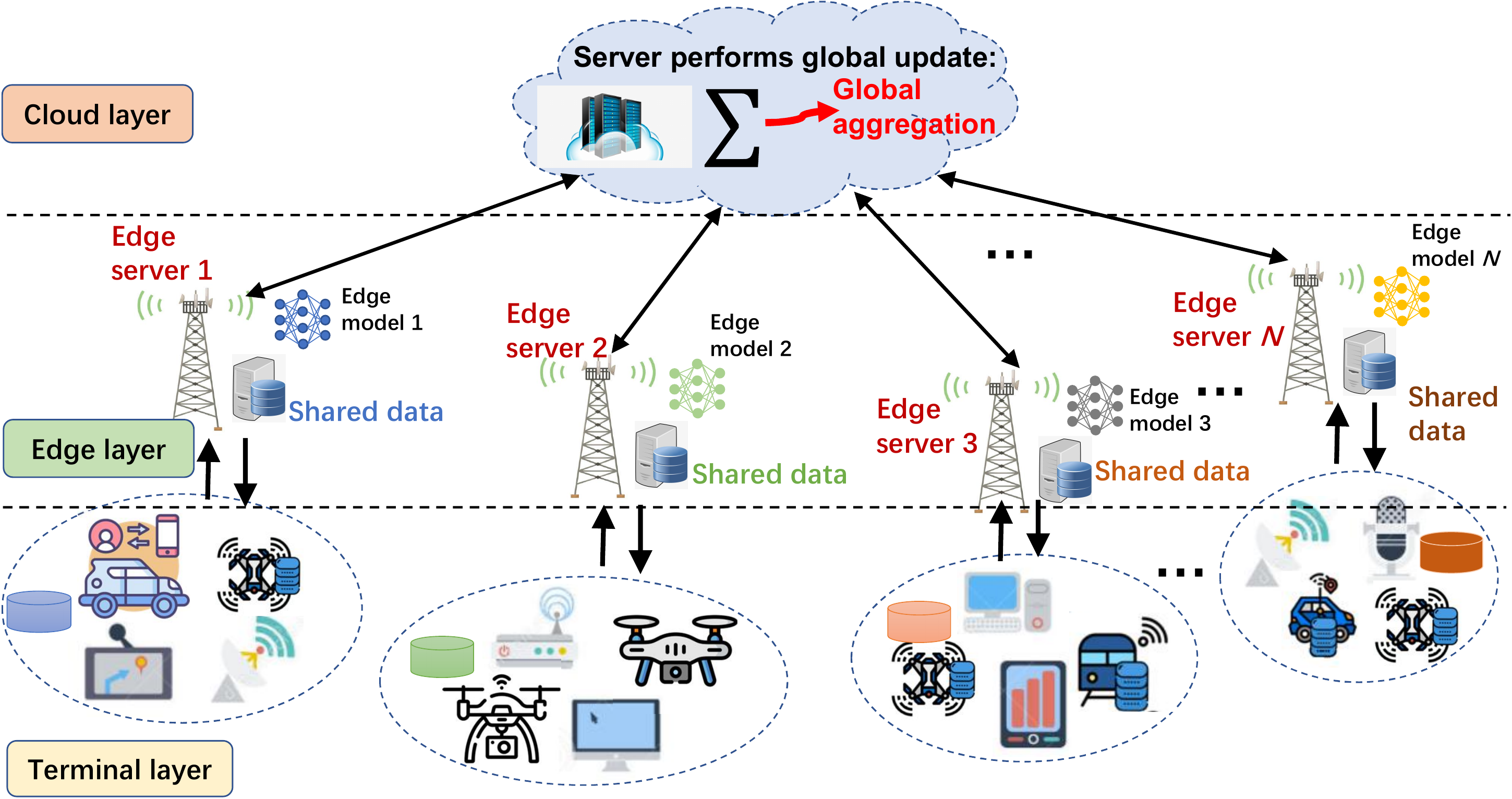}
			\label{fig:har_s50a}
		\end{subfigure}\\%\\
		\subcaption{Hierachical FL architecture.}
	\end{minipage} \hfill

	\begin{minipage}[t]{0.95\linewidth} % 如果一行放2个图，用0.5，如果3个图，用0.33
		\centering
		\begin{subfigure}{1\textwidth}
			\centering
			\includegraphics[width=1.0\linewidth]{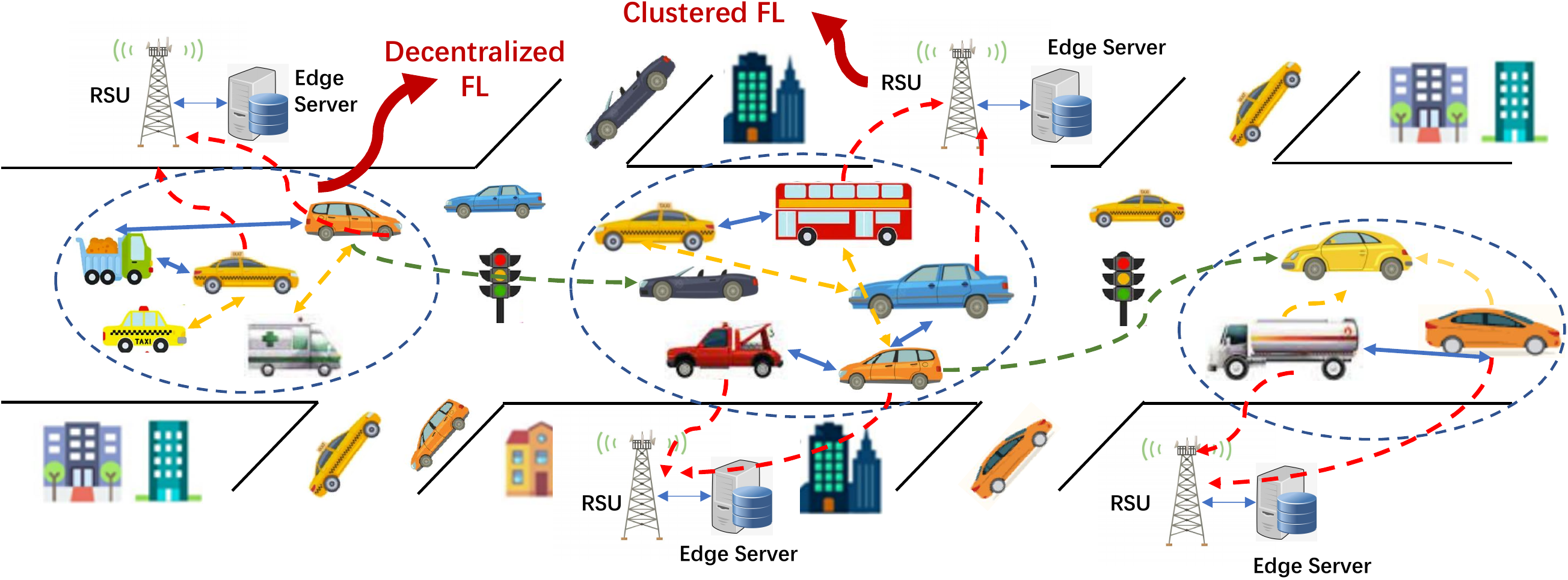}
		\end{subfigure}\\%\\
		\subcaption{Decentralized and clustered FL architectures.}
	\end{minipage} 	
	\caption{Illustration of existing FL architectures, including centralized, hierachical, decentralized and clustered FL architectures. It can be seen these architectures are generally existing in hybrid mode to handling networks with complicated topology.}
	\label{fig:fedarch}
\end{figure*}

%\begin{figure*}
%	\centering
%	\begin{minipage}[t]{0.48\linewidth}
%		\begin{subfigure}
%			\centering
%			\includegraphics[width=1.0\linewidth]{figures/cenfed}
%		\end{subfigure}\\			
%		%\caption{fig1}
%	\end{minipage} \hfill 
%	
%	\begin{minipage}[t]{0.5\linewidth}
%		\begin{subfigure}
%			\centering
%			\includegraphics[width=1.0\linewidth]{figures/hirfed}
%		\end{subfigure}		\\	
%		%\caption{Hierachical FL architecture.}
%	\end{minipage} \hfill 
%	
%	\begin{minipage}[t]{0.9\linewidth}
%		\begin{subfigure}
%			\centering
%			\includegraphics[width=1.0\linewidth]{figures/decenfed}
%		\end{subfigure}			
%		%\caption{Decentralized and clustered FL architectures.}
%	\end{minipage}%
%	
%	\centering
%	\caption{ Illustration of existing FL architectures, including centralized, hierachical, decentralized and clustered FL architectures. It can be seen these architectures are generally existing in hybrid mode to handling networks with complicated topology.}
%	\label{fig:fedarch}
%\end{figure*}

\para{Decentralized Intra-cluster and Centralized Inter-cluster.}In this architecture, clients in each cluster cooperatively learn the local models via device-to-device (D2D) communications in a decentralized manner, while  the central server coordinates the clusters in the centralized fashion. Since client privacy cannot be decrypted by the central server, the privacy is highly improved. Moreover, by clustering the homeomorphic clients together, this scenario is able to alleviate the heterogeneous issue.  Based on the architecture, Lin et al. \cite{DFL_Lin2021} proposed two timescale hybrid FL, which outperforms the current state-of-the-art counterparts in terms of the model accuracy and network energy consumption. To guarantee the clustered structure, we can utilize an adaptive strategy without manual intervention to realize dynamic clustering \cite{CFL_Gong2022}.

\subsection{Summarization}
Centralized FL is fundamental in this field, but due to the simple and fixed pattern, it is not flexible and scalable, which will significantly limit the applications in large-scale networks. Decentralized FL can reach the learning tasks by only requiring the clients communication with their neighbors. We can expect it will have great potential for applications in large-scale networks. However, it still demands high performance algorithms and insurance in privacy and security. Compared with centralized FL and decentralized FL, the clustered FL is more adaptable to different scenes, and can better balance performance and privacy security, we can expect its significant potentiality in the complicated systems, e.g., large-scale dynamic IoV. 

	\section{Federated Base Model}\label{apa}
While FL techniques have laid the foundation of massive applications, the fast growing applications will adversely demand the development of techniques. Through comprehensively surveying the applications of FL, we further deeply abstract the application problems and explore the underlying techniques that support these applications.   As a result, \textbf{they can be generally categorized into a basic paradigm consisting of a federated base model combined with a federated training framework}. Specifically, we first  discuss several popular machine learning models, which can be straightforwardly trained through the popular FL methods. Furthermore, we discuss the nonsmooth regularization due to its promising potential in biomedical applications. Considering the more general and practical fashion that different clients may have different tasks, we introduce multi-task learning under FL solution. Next follows matrix factorization, which has been widely employed in recommendation systems, our strategy is to formulate the problem with nuclear norm to accurately constrain the matrix being low-ranked.  In general, our goal is to provide the FL schemes  covering sufficient cases to facilitate larger-scale FL applications.
\scriptsize
\begin{table*}[ht]
	\renewcommand{\arraystretch}{1.3}
	\setlength{\abovecaptionskip}{0.cm}
	\setlength{\belowcaptionskip}{-0.cm}
	\begin{center}
		\begin{tabular}{l|l}
			\hline
			\textbf{Model}& \textbf{Federated training procedure}                  \\ \hline
			MLP& \multirow{4}{*}{\tabincell{l}{The client performs multiple SGD to obtain the local model, \\and then the server performs aggregation to update the global model.}} \\ \cline{1-1}
			CNN&                   \\ \cline{1-1}
			LSTM&                   \\ \cline{1-1}
			GRU&                   \\ \cline{1-1}\hline
			TM&\tabincell{l}{The client performs Metropolis Hastings based private computations to obtain the local model, \\then the server performs  integration of the received local topic models.}                   \\ \hline
			\tabincell{l}{Nonsmooth\\ regularization}& \tabincell{l}{The client performs multiple mirror descent steps to update the local model and \\then the server performs aggregation to update the global model. }                \\ \hline
			MTL& \tabincell{l}{The client performs ADMM subproblems to update the local model and Lagrangian multiplier \\and the server performs the aggregation to obtain the global model corresponding to the task.}                   \\ \hline
			MF&  \tabincell{l}{The client performs ADMM subproblems to update user item vector and Lagrangian multiplier\\ and the server performs the subproblem to update the item profile matrix. }                \\ \hline
		\end{tabular}
	\end{center}
	\caption{Training models via FL paradigm.}
\end{table*}
\normalsize

\subsection{Machine Learning Models}
Many machine learning models can be collaboratively trained via FL framework, e.g., as the representatives models in deep learning, the multi-layer perceptron (MLP) and convolutional neural network (CNN)  have been widely used. Specifically, with the same structures of these models on each client, they can be locally trained with multiple steps by utilizing their own datasets, then the updated local model is transmitted to the server for further aggregation (such as the weight averaging) \cite{pmlr-v54-mcmahan17a}.  Another popular model is the long short-term memory (LSTM), which has been utilized in FL for  the language modeling \cite{pmlr-v54-mcmahan17a}. Gated recurrent unit neural network (GRU) based on FedAvg solution has been applyed for the traffic flow prediction \cite{9082655}.  In summary, we can train these models which share the common procedure of multiple local updates and global aggregation.  Another popular machine learning model - topic modeling,  has been applied in many industrial fields, such as medical informatics \cite{DBLP:conf/sigir/ArnoldS12}, and web search log mining \cite{DBLP:conf/cikm/HarveyCC13}. It has been extended to FL settings  due to the increasing concerns on the data privacy issue \cite{DBLP:conf/cikm/JiangSTWZXY19,DBLP:journals/tist/JiangTSWZPLXY21}. Specifically, Jiang et al. \cite{DBLP:conf/cikm/JiangSTWZXY19} propose a federeted topic modeling (FTM) framework to discover topics in a collection of documents and classify a document within the collection over distributed massive clients. Moreover, the local clients perform  Metropolis Hastings based private computations to obtain the local model, then the server implements  integration of the received local topic models. FTM has been demonstrated to simultaneously maintain the privacy and improve the training quality.
 
\subsection{Nonsmooth Regularization}
Nonsmooth regularization has been widely applied in the biomedical applications with the sparse learning for dealing the challenges of large dimension data, redundant features and  sample absence \cite{6165264,MTL_FED}. Moreover, the biomedical application data is high sensitive. Therefore,  FL can be a promising technique for the cooperation over medical organizations, while it helps to reach the high accuracy solution, the privacy and security is preserved  without data centralization. To start with, we assume there are $N$ medical organizations and denote $\mathcal{D}_i$ as the local dataset on the $i$th medical organizations. Subsequently,  the federated problem can be formulated in the following:
\begin{equation}
	\underset{w\in\mathbb{R}^d}{\text{min  }} \frac{1}{N}\sum_{i=1}^{N}f_i(w)+\alpha\Psi(w)
\end{equation}
where $w\in\mathbb{R}^d$ is the model, $f_i(w):\mathbb{R}^d\rightarrow\mathbb{R}$ is the local loss function, and $\Psi:\mathbb{R}^d\rightarrow \mathbb{R}$ is the nonsmooth convex regularizer. When  $f_i(w) = \frac{1}{n_i}\sum_{(x_i,y_i)\in\mathcal{D}_i}\norm{w^Tx_i-y_i}^2$, with $\Psi=\norm{\cdot}_1$, the problem is known as LASSO. 

We next discuss the FL solution over the $N$ medical organizations via FedMirror, where the local update utilizes the multiple mirror descent steps \cite{DBLP:conf/icml/YuanZR21}. Specifically, we introduce the Bregman distance. Let $h:\mathbb{R}^d\rightarrow\mathbb{R}$ be a continuously-differentiable, strictly convex function, then the Bregman distance can be defined: $\mathcal{D}_h(x,y)=h(x)-h(y)-\langle\nabla h(y),x-y\rangle.$
Next, the local model $w_i$ is mapped to the dual space to form the dual model $z_i$ with the carrier $\nabla h$, i.e., $z_i\leftarrow\nabla h(w_i)$. Subsequently, the dual model $z_i$ is updated by utilizing the local gradient queried at $w_i$, i.e., $z_i\leftarrow z_i-\eta\cdot \nabla f_i(w_i)$. Finally, the updated dual model is mapped back to the primal space, i.e., $w_i\leftarrow\nabla (h+\eta\Psi)^*(w_i)$, where for any convex function $g$, $g^*(z)$ is the convex conjugate function given by $g^*(z)=\text{sup}_{w\in\mathbb{R}^d}\big \{\langle z,w\rangle-g(w)\big\}$. With multiple local mirror descent steps, the model $w_i$ is transmitted to the server for further aggregation, namely $w\leftarrow{\nicefrac{1}{N}\sum_{i=1}^{N}w_i}$.

%\subsubsection{FedDualAvg} However, FedMirror suffers from the ``Curse of Primal Averaging'', i.e., while the local model $w_i$ is sparse after mapping back to the primal space,   the global model returns to dense vector when server performs aggregation. To mitigate this problem, FedDualAvg is proposed \cite{DBLP:conf/icml/YuanZR21}. Different with FedMirror that alternates the dual and primal space in the local updates, FedDualAvg always operates in the dual space in both the local updates and the global aggregation, maintaining the consistency and mitigating the dense aggregation on the server. Specifically, it first initializes the local dual model $z_i$ on each medical organization $i$, then updated by the local gradient queried at the primal model. After the local updates, each local dual model $z_i$ is transmitted to the server for dual averaging, i.e., $z\leftarrow{\nicefrac{1}{N}\sum_{i=1}^{N}z_i}$. It should be noted only after the algorithm converges, the medical organizations can obtain the global model by  mapping back with the dual global model $z$ via the carrier $\nabla (h+\beta\Psi)^*(z)$.

\subsection{Low Rank Matrix Estimation}
Low rank matrix estimation (LRME) is an efficient technique in dealing with large-scale problems with high-dimensional data and thus rising the significant interests in scientific computing. Exemplar applications include the sparse principle component analysis \cite{pmlr-v28-papailiopoulos13}, data compression for natural language processing  \cite{NEURIPS2021_f56de5ef} and  distance matrix completion \cite{6160810}. In this section, we illustrate LRME in federated settings. In particular, we propose to solve the problem via linearized federated ADMM (L-FedADMM), since its base algoritm ADMM is a convenient and efficient tool for solving nonsmooth optimization problems \cite{DBLP:journals/ftml/BoydPCPE11}. 
\begin{algorithm}[ht]
	\renewcommand{\algorithmicrequire}{\textbf{Input:}}
	\renewcommand{\algorithmicensure}{\textbf{Output:}}
	\caption{L-FedAdmm} 
	\label{alg:fedadmm}
	\begin{algorithmic}[1]
		\STATE \textbf{server input:}  initial $Z^0$, $\eta_g$.
		%\ENSURE $y = x^n$ 
		\STATE \textbf{client $i$'s input:} initial $X^0_i$, $\pi_i^0$ and $\eta_l$.
		
		\FOR{$r=1,\dots,R$}
		\STATE 
		
		{\textbf{Server implements} steps 5-6:}
		
		\STATE
		Obtain the update $Z^r$ by performing the iteration (\ref{Proxzz})
		
		\STATE 
		
		Sample clients ${\mathcal{S}}\subseteq[N]$ and	transmit $Z^{r}$ to client $i\in{\mathcal{S}}$.
		
		\STATE
		{\textbf{Clients implement} steps 8-14 \textbf{in parallel for} $i\in{\mathcal{S}}$:}
		
		\STATE
		After receiving $Z^r$, client $i\in{\mathcal{S}}$ performs
		
		\FOR{$t=1,\dots,T$}
		\STATE 
		Obtain $X_i^t$ via (\ref{X}).
		\STATE 
		Update $\pi_i^t$: $\pi^t_i\leftarrow \pi^{t-1}_i+\rho(X^t_i-Z^r)$.
		
		\ENDFOR
		\STATE Set $X^r_i\leftarrow X^T_i$ and $\pi^r_i\leftarrow \pi^T_i$ for $i\in\mathcal{S}$, and $X^{r}_i\leftarrow X^{r-1}_i$ and $\pi^{r}_i\leftarrow \pi^{r-1}_i$ for $i\notin\mathcal{S}$.
		\STATE Client $i$ transmits $\pi_i^r+\rho X^r_i$ to the server.

		\ENDFOR
	\end{algorithmic}
\end{algorithm}
For the adaptation to FedOpt, we first formulated the federated LRME problem. Assume there are $N$ distributed clients, and denote $D_j\in\mathbb{R}^{d\times d}$ as the data sample.  Then  the local loss function of LRME on client $i$ is defined as  $f_i(X)=\sum_{j=1}^{n_i}(\langle X,D_j\rangle-y_j)^2$, where $y_j$ is the noisy observation. All clients target at solving the following problem via L-FedADMM:
\begin{equation}\label{lowrankestimation}
	\underset{X}{\text{min }} \sum_{i=1}^{N} f_i(X) + \lambda\norm{X}_*,
\end{equation}
Here, we use the nuclear norm $\norm{\cdot}_*$ to constrain the  low rank of the matrix.  The problem  (\ref{lowrankestimation}) targets at recovering the low rank matrix $X\in\mathbb{R}^{d\times d}$ from the noisy observations $y_j$. For L-FedADMM, the consensus optimization which is equivalent to (\ref{lowrankestimation}) is derived: 
\begin{equation}\label{consensus}
	\begin{aligned}
		&\underset{X}{\text{min }} \sum_{i=1}^{N} f_i(X_i) + \lambda\norm{Z}_*, \quad\text{s.t. } X_i-Z = 0, i=1,\dots,N,
	\end{aligned}
\end{equation}
where $Z\in\mathbb{R}^{d\times d}$ is the consensus variable, $X_i\in\mathbb{R}^{d\times d}$ can be viewed as the local model. Obviously, (\ref{consensus}) and (\ref{lowrankestimation}) are equivalent.  Then, the augmented Lagrangian function can be further derived \cite{DBLP:journals/ftml/BoydPCPE11}: 
\begin{equation}\label{lagrangian}
	\begin{aligned}
		\mathcal{L}(X_{1:N}, \Pi, Z ) = \sum_{i=1}^{N} &\big\{f_i(X_i) + \left\langle \pi_i, X_i-Z \right\rangle + \frac{\rho}{2}\norm{X_i-Z}^2_2\big\} + \lambda\norm{Z}_*,
	\end{aligned}
\end{equation}
where $\pi_i\in\mathbb{R}^{d\times d}$ is the Lagrangian multiplier, $\left\langle , \right\rangle$ denotes the matrix inner product, $\rho>0$ is the regularization parameter and $\Pi=\{\pi_i,i=1,\dots,N\}$. Subsequently, the decentralized ADMM \cite{DBLP:journals/ftml/BoydPCPE11} solves the problem (\ref{consensus}) by alternating the update. To be specific, the client $i$ download the consensus matrix $Z$ and perform the update via minimizing (\ref{lagrangian}), i.e.,  $X^+_i\leftarrow{\text{argmin}_{X_i}}\mathcal{L}(X_{1:N}, \Pi, Z )$, then it continues to update the Lagrangian multiplier: $\pi^+_i\leftarrow \pi_i+\rho(X^+_i-Z)$. Finally, with $X_i$ and $\pi_i$, the server update the consensus matrix $Z$ via $Z^+\leftarrow{\text{argmin}_Z}\mathcal{L}(X^+_{1:N}, \Pi^+, Z )$.

We further expand for obtaining the update $X_i^+$ on client $i$. To be specific,  the efficient linearized approximation strategy for $f_i(X_i^+)$ at $X_i$ can be utilized, namely $f_i(X_i^+)\approx f_i(X_i)+\langle \nabla f_i(X_i),X_i^+-X_i\rangle+\nicefrac{1}{2\eta_l}\norm{X_i^+-X_i}^2_2$, where $\eta_l$ is the second-order approximate. By expanding, the closed-form solution can be efficiently derived via:
\begin{equation}\label{X}
	X_i^+\leftarrow \frac{\rho\eta_lZ-\eta_l\pi_i+X_i-\eta_l\nabla f_i(X_i)}{1+\rho\eta_l}.
\end{equation}
Furthermore, with the updated $X_i^+$ and $\pi^+_i$, we further derive the update for $Z$. Specifically, the proximal operator for the nuclear norm iteratively is employed: 
\begin{equation}\label{Proxzz}
	\begin{aligned}
		Z^{i+1}\leftarrow  \text{ Prox}_{\lambda\eta_g\norm{\cdot}_*} &\big\{Z^i-\eta_g\big(N\rho Z^{i}-\sum_{i=1}^{N}\{\pi^+_i+\rho X_i^+\}\big)\big\}
	\end{aligned}
\end{equation}
for $i=0,\dots,I-1$, then we can obtain $Z^+\leftarrow Z^M$. Next, we mimic the adaptation of SGD to FedAvg for L-FedADMM \cite{pmlr-v54-mcmahan17a}. Precisely, the participated clients $i\in\mathcal{S}$ performs the local updates for $X_i$ (via (\ref{X})) and $\pi_i$ for multiple times (say $T$ local iterations), and then the quantity $\pi_i^{+}+\rho X^{+}_i$ is transmitted to the server for performing the global update (\ref{Proxzz}). While clients $i\in \mathcal{S}$ update their  local models $X_i$ and Lagrangian multipliers $\pi_i$, those clients $i\notin\mathcal{S}$ hold their $X_i$ and  $\pi_i$,  i.e., $X^{+}_i\leftarrow X_i$ and $\pi^{+}_i\leftarrow \pi_i$.  We summarize L-FedADMM in Algorithm \ref{alg:fedadmm}.

%To be specific, let the words $w$ be the training data on one client and $z$ be the latent topics assigned to $w$. Subsequently, under the general assumption of LDA \cite{DBLP:journals/jmlr/BleiNJ03}, the joint probability can be derived as:
%\begin{equation}
%	P(z_{di}|z_{-di},w,\alpha,\beta)\propto\frac{C^{DK}_{dz_{di}}+\alpha}{\sum_{k'}(C^{DK}_{dk'}+\alpha)}P(w_{di}|z_{di},\Phi),
%\end{equation}
%where $P(w_{di}|z_{di},\Phi)$ is in the exponential family. 

\subsection{Multi-Task Learning}
In real-world applications, there is an urgent need solving the problem of multiple learning tasks simultaneously, while exploiting commonalities and differences across tasks. As a running example, consider learning the behaviors of drivers on the road in the transport network based on their individual radar signal, image and trajectory. Each driver may generate data through a  distinct distribution, and as a consequence,  it is natural to fit the multiple distinct models to the decentralized data.   This has resulted in the technique of multi-task learning (MTL), where  some clients have the shared task, while other clients have different ones. Since MTL learns separate models for each task among the clients,  FL is naturally suited to MTL \cite{DBLP:conf/nips/SmithCST17}. Next follows the formulated federated MTL problem, existing solution  and  our proposed approach.

In general, suppose there are $N$ clients with $m$ tasks, and  each client $i$ corresponds to the task $\psi(i)\in\{1,\dots,m\}$. For client $i$, it targets at learning the model $z_{\phi(i)}\in\mathbb{R}^d$ to fit its own data. Therefore, the federated MTL aims to learn the joint model $Z=\{z_1,\dots,z_m\}$ over the $N$ clients and the federated problem is formulated as follows: 

\begin{equation}
	\begin{aligned}\label{mtl2}
		& \underset{Z}{\text{min}}
		& & \sum\nolimits_{i=1}^{N}f_i(z_{\phi(i)})+R(Z)
	\end{aligned}
\end{equation}
where $f_i(\cdot):\mathbb{R}^d\rightarrow\mathbb{R}$ is the local loss function, and $R:\mathbb{R}^d\rightarrow \mathbb{R}$ is the regularization function. MOCHA solves the problem (\ref{mtl2}) via its dual problem with convex conjugate of the objective function \cite{DBLP:conf/nips/SmithCST17}. For the local step in MOCHA, it minimizes  a quadratic approximation of the dual subproblem. However, the approach suffers from the strong assumption, e.g., MOCHA assumes one-to-one mapping, namely each client $i$ corresponds to the task $i$. Moreover, MOCHA requires all clients participating in the update at each iteration, which is unrealistic in many applications. 

\begin{figure}
	\centering
	\includegraphics[width=0.8\linewidth]{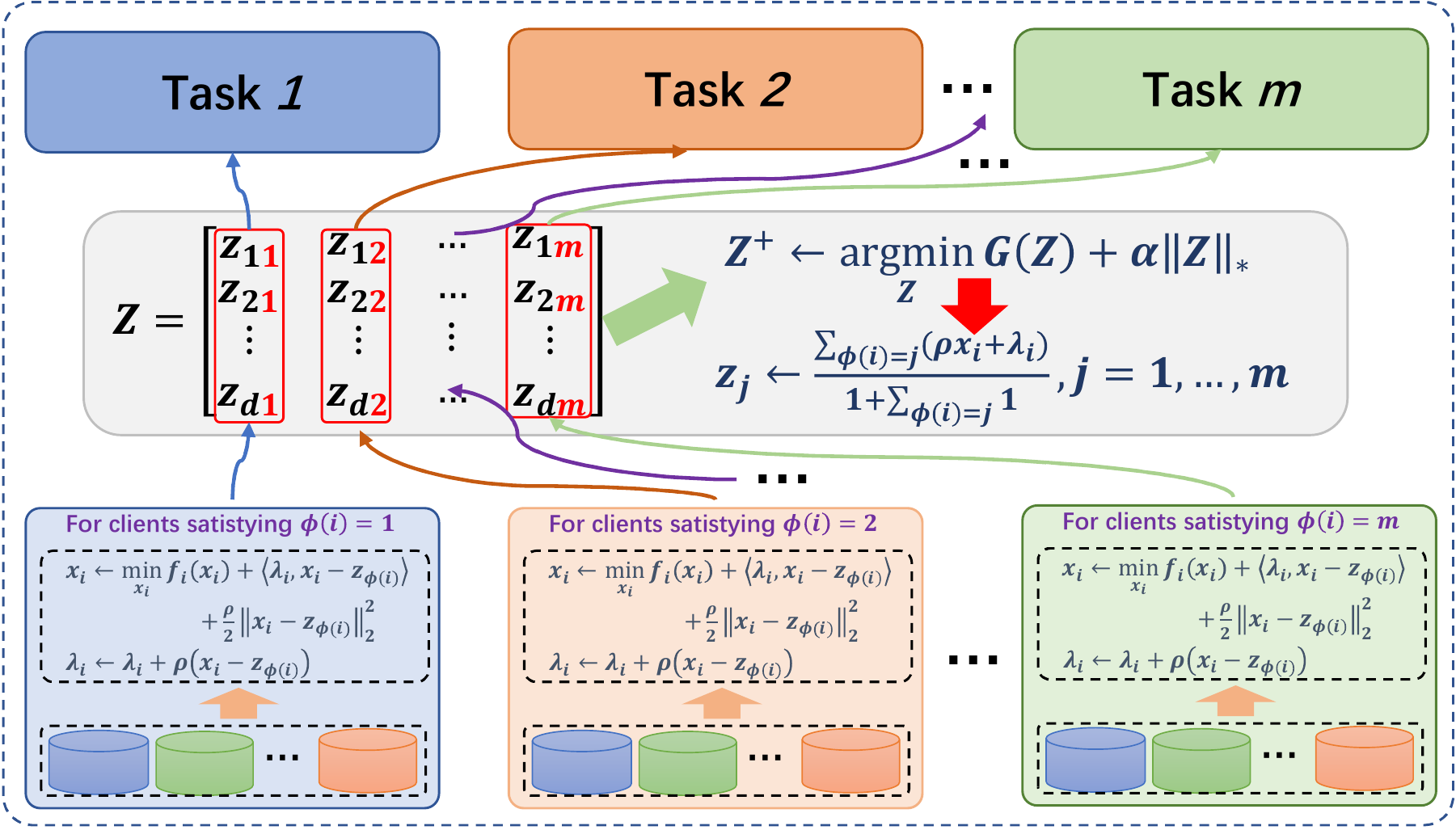}
	\caption{Multi-task learning via FL with ADMM strategy, the basic idea is each task corresponding to multiple clients, namely, for task $j, j=1\dots,m$, its corresponding clients satisfy $\psi(i)=j$, where $i=1,\dots,N$. }
	\label{fig:fedmtl}
\end{figure}

\para{Decentralized ADMM for MTL.} Since the existing approach MOCHA has limitations for wide applications, we propose a novel strategy  based on ADMM to handle MTL. Our proposed approach exactly fits the centralized FL architecture. Moreover, our proposed approach is simple and convenient and thus is promising to be deployed in much wider applications.   

To be specific, we  view the local model $z_i$ as the consensus variable, and introduce the variable $x_i\in\mathbb{R}^d$ as the new local model for training. Thus the MTL model that $N$ clients aim to collaboratively learn is $\{x_i,\,i=1,\dots,N\}$.  Subsequently, we formulate the federated consensus optimization problem over the $N$ clients as
\begin{equation}
\begin{aligned}\label{mtl}
& \underset{x_{i},i=0,\dots,N;}{\text{min}}
& & \sum\nolimits_{i=1}^{N}f_i(x_i)+R(Z), \quad\quad\text{ s.t.}
%& & f_i(x) \leq b_i\\
& & x_i=z_{\psi(i)}, i=1,\dots,N
\end{aligned}
\end{equation}
Then, we will use the distributed ADMM discussed in Section \ref{opt} to solve the above federated problem. For the local model $x_i$ update, it can be obtained via standard subproblem minimization in ADMM:
\begin{equation}\label{x_i}
	\begin{aligned}
		\underset{x_i}{\text{min }}f_i(x_i)+&\left\langle \lambda_i,x_i-z_{\phi(i)}\right\rangle+\frac{\rho}{2}\norm{x_i-z_{\phi(i)}}^2_2.
	\end{aligned}
\end{equation}
where $\lambda_i$ is the Lagrangian multiplier. In addition to local model update, the client also needs to update  $\lambda_i$: $\lambda_i\leftarrow\lambda_i+\rho(x_i-z_{\phi(i)})$. It should be noted that while (\ref{x_i}) can be solved via standard SGD, linear approach can be adopted for efficient computation.  In the sequel, we discuss to solve the consensus matrix $Z$ on the server by considering three scenarios:  
\begin{itemize}[leftmargin=*]
	\item First, consider the case when $\psi(i)=i$ and $R(Z)=\alpha\norm{Z}_*$, namely we choose the nuclear norm to be the regularization function to exploit the commonalities and differences across tasks and obtain a low rank matrix solution $Z^*$. Moreover, $\psi(i)=i$ means there are $N$ tasks and each task $i$ corresponds to each client $i$ (the same assumption with MOCHA). Then the updated consensus matrix $Z^+$ with low rank can be derived by solving the following subproblem:
	\begin{equation}\label{Z+}
		Z^+\leftarrow \underset{Z}{\text{ argmin } }\alpha\norm{Z}_*+\frac{\rho}{2}\norm{X-Z+\frac{1}{\rho}\Lambda}^2_F,
	\end{equation}
	where $X=[x_1,\dots,x_n]$, $Z=[z_1,\dots,z_n]$ and $\Lambda=[\lambda_1,\dots,\lambda_n]$. Then, we can acquire the closed-form solution $Z^+\leftarrow U\cdot\text{diag}\big\{\text{prox}_{\frac{\alpha}{\rho}\norm{\cdot}_1}(\sigma)\big\}\cdot V^T$, where $U$, $\sigma$ and $V$ can be conveniently obtained via the singular value decomposition of the matrix $(X+\nicefrac{1}{\rho}\Lambda)$.
	\item Second, consider the scenario of  $\psi(i)\neq i$ and $R(Z)=\alpha\norm{Z}_*$, then the subproblem is formulated in the following:
	\begin{equation}\label{subproblem}
		\begin{aligned}
			Z^+\leftarrow\underset{Z}{\text{ argmin } }G(Z)+\alpha\norm{Z}_*,
		\end{aligned}
	\end{equation}
	where for simplicity we denote $G(Z)$ as follows:
	\begin{equation}\label{GZ}
		G(Z):=\sum_{i=1}^{N}\left(\left\langle \lambda_i,x_i-z_{\phi(i)}\right\rangle+\frac{\rho}{2}\norm{x_i-z_{\phi(i)}}^2_2\right).
	\end{equation}
	It can be seen the subproblem (\ref{subproblem}) has no closed-form solution, therefore, we propose to solve  (\ref{subproblem}) via proximal gradient descent.  Specifically, the gradient $\nabla_Z G(Z)$ can be derived column by column, i.e., $\nabla_{z_j} G(Z), j=1,\cdots,m$ as follows $\nabla_{z_j} G(Z)=\sum_{\phi(i)=j}\rho z_j-(\rho x_i+\lambda_i)$. 
	Subsequently, with the gradient $\nabla_Z G(Z)$, we are able to solve the subproblem (\ref{subproblem}) via the iteration $i=0,I-1$ 
	\begin{equation}\label{Proxz}
		Z^{i+1}\leftarrow  \text{ Prox}_{\lambda\eta_g\norm{\cdot}_*} \left\{Z^i-\eta_g\nabla_Z G(Z^i)\right\},
	\end{equation}
	where $\eta_g>0$ is the step size.
	\item Third, consider the case when  $\psi(i)\neq i$ and the regularization function $R(Z)=Tr(ZZ^T)$, then the consensus variable $Z$ can be obtained via the closed-form solution:
	\begin{equation}\label{closed-form}
		z_j\leftarrow\frac{\sum_{\psi(i)=j}(\rho x_i+\lambda_i)}{1+\rho\sum_{\psi(i)=j}1}, j=1,\cdots,m.
	\end{equation}
\end{itemize}

\begin{algorithm}[ht]
	\renewcommand{\algorithmicrequire}{\textbf{Input:}}
	\renewcommand{\algorithmicensure}{\textbf{Output:}}
	\caption{Decentralized ADMM for MTL} 
	\label{alg:dADMM-MTL}
	\begin{algorithmic}[1]
		\STATE \textbf{server input:} The communication round $R$ and iteration number $I$, initial $Z$, $\alpha$ and $\rho$.
		%\ENSURE $y = x^n$ 
		\STATE \textbf{client $i$'s input:} The local iteration number $T$, initial $\eta$, $x_i$, and $y_{i,j}$. 		
		\FOR{$r=1,\dots,R$}
		\STATE 
		
		{\textbf{Server implements} steps 5-6:}
		\STATE
		Updates $Z$ considering the three scenarios respectively,
		\begin{itemize}
			\item if $\psi(i)=i$ and $R(Z)=\alpha\norm{Z}_*$:
			
			Obtain $Z^+$ via (\ref{Z+}).
			\item if $\psi(i)\neq i$ and $R(Z)=\alpha\norm{Z}_*$:
			
			Perform (\ref{Proxz}) via the iteration $i=0,I-1$ 
			\item if $\psi(i)\neq i$ and $R(Z)=Tr(ZZ^T)$:
			
			Update $Z$ via (\ref{closed-form}).
		\end{itemize}

		\STATE 
		
		Randomly sample clients ${\mathcal{S}}\subseteq[N]$ and	transmit $z_{\phi(i)}$ to each client $i\in{\mathcal{S}}$.
		
		\STATE
		{\textbf{Clients implement} steps 8-10 \textbf{in parallel for} $i\in{\mathcal{S}}$:}
		
		\STATE
		After receiving $z_{\phi(i)}$, client $i\in\mathcal{S}$ updates $x_i$ by solving the subproblem (\ref{x_i}), and
		\STATE updates  $\lambda_i$: $\lambda_i\leftarrow\lambda_i+\rho(x_i-z_{\phi(i)})$
		\STATE Client $i$ transmits $(\rho x_i+\lambda_i)$ to the server.		
		\ENDFOR
	\end{algorithmic}
\end{algorithm}

\subsection{Matrix Factorization}
Matrix factorization (MF) is a prominent machine learning technique which has widely applied in various domains including recommendation system \cite{DBLP:journals/tissec/KimKKYSK18,5197422} and environment monitoring \cite{7509395}. While MF has been developed with security guarantee in centralized scenarios \cite{8241854,DBLP:journals/tissec/KimKKYSK18}, its application in FL scenarios is still under investigation.   Chai et al. \cite{9162459} propose a secure federated MF. However, it is susceptible to the error data and has not fully exploited the item correlations. Hence, we  propose a robust and efficient strategy for the secure   federated recommendation. 

\begin{figure}
	\centering
	\includegraphics[width=0.7\linewidth]{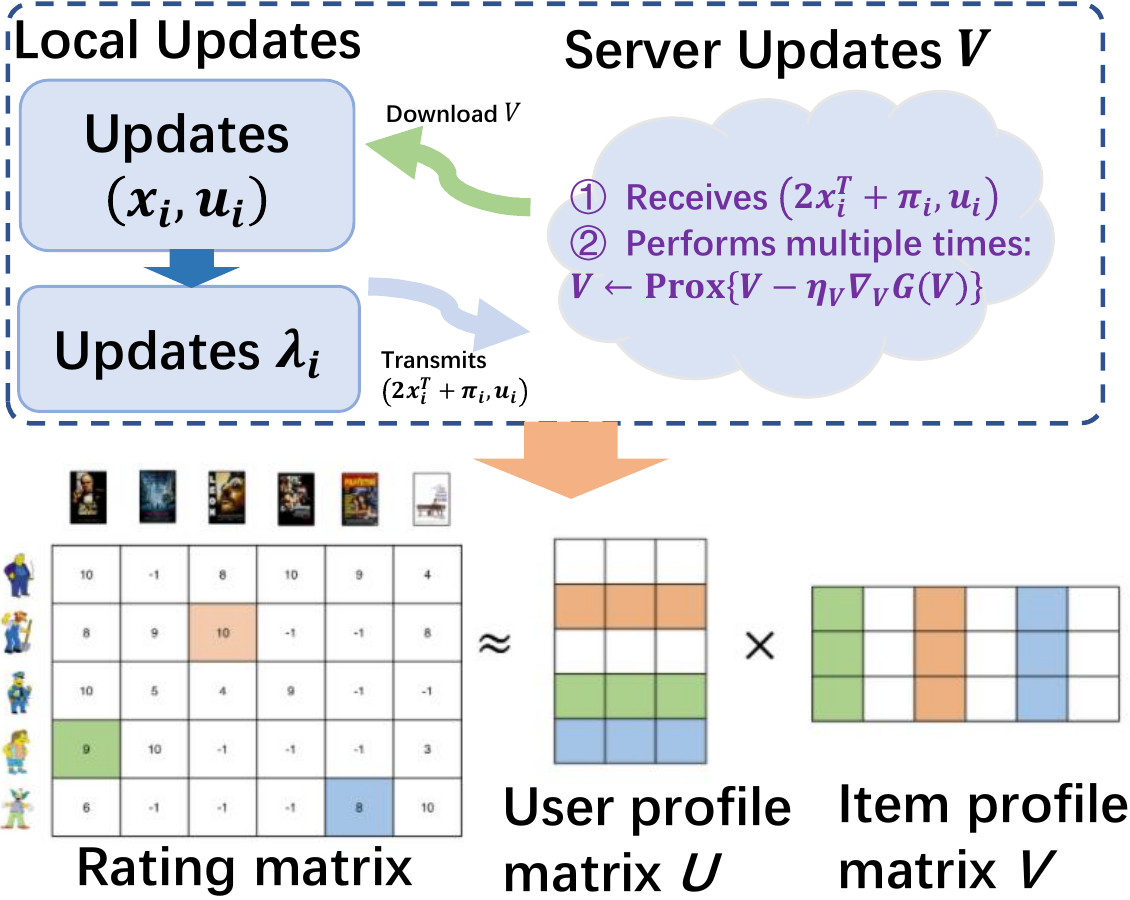}
	\caption{Federated matrix factorization. Its main ingredient is: first, each user holds there rating vector $R_{i,:}$, moreover, it initializes a user profile vector $u_i$, $(R_{i,:},u_i)$ are kept private. The item matrix $V$ can be shared through users. They collaborate with each other to learn $(U,V)$.}
	\label{fig:fedmf}
\end{figure}

In the sequel, we first provide the MF model.  Suppose there are $N$ users and each user  has rated a subset of $M$ items, which forms the rating matrix $R\in\mathbb{R}^{N\times M}$, with the element $r_{i,j}$ representing the $i$th user rating the $j$th item. Moreover, we use $R_{i,:}$ to denote the $i$th row of the matrix $R$, which also means the $i$th user's rating on all the $M$ items. Then, the recommendation system targets at predicting all users' rating on all items. To adopt the MF technique, the problem can be formulated as the decomposition of the rating matrix into the user profile matrix $U\in\mathbb{R}^{N\times d}$ and  item profile matrix $V\in\mathbb{R}^{M\times d}$. Subsequently, the prediction of the $i$th user rating on the $j$th item can be given as  $\langle u_i,v_j\rangle$, where $u_i$ and $v_j$ are the $i$th and $j$th row vectors in $U$ and $V$, respectively. To learn $U$ and $V$ in federated settings, the following nonsmooth regularization problem is formulated:
\begin{equation}
	\underset{U,V}{\text{min } } \sum_{i=1}^{N} f_i(u_i,V)+\lambda \norm{U}_{2,1} +\mu\norm{V}_*,
\end{equation} 
where $f_i(u_i,V)=\norm{R_{i,:}-u_i\cdot V^T}^2_2$. Here, we use the nuclear norm and $L_{2,1}$ norm instead of the 2-norm  in \cite{9162459} to make the our approach exploit well in the item dependency and more robust to the data error \cite{10.1145/1143844.1143880}. For the FedOpt solution, we propose a three-blocked distributed ADMM based framework.  First of all, we incorporate a concensus matrix $X$, with its $i$th row $x_i=u_i\cdot V^T$, subsequently the above MF problem can equivalent reformulated in the following three-blocked consensus problem with respect to $(X,U,V)$:
\begin{equation}
\begin{aligned}\label{MF_admm}
& \underset{X,U,V}{\text{min}}
& & \sum\nolimits_{i=1}^{N}f_i(x_i)+\lambda \norm{U}_{2,1} +\mu\norm{V}_*,  \text{ s.t.}
%& & f_i(x) \leq b_i\\
& & x_i-u_i\cdot V^T=0, i=1,\dots,N.
\end{aligned}
\end{equation}
where $f_i(x_i)=\norm{x_i-R_{i,:}}_2^2$. With distributed ADMM, we propose the following FedAvg-like architecture approach, namely the item profile matrix $V$ is updated on the server side while the $i$th row of the consensus matrix $X$ and its own user profile vector $u_i$. Moreover,  the Lagrangian multiplier $\pi_i$ is also updated  on the client $i$. 

To be specific, after the sever receives $(2x^T_i+\pi_i,u_i)$, it updates $V$  on the server via the minimization:
\begin{equation}\label{V}
	V\leftarrow \underset{V}{\text{argmin }} \mu\norm{V}_*+G_V(V).
\end{equation}
where  for simplicity we denote the summation term as $G_V(V):=\sum_{i=1}^{N}\norm{x_i-u_iV^T}^2_2+\sum_{i=1}^{N}\langle \pi_i,-u_iV^T\rangle$. Subsequently, we aim to solve the problem (\ref{V}) via the proximal operation. We illustrate this by first deriving of the gradient $\nabla_V G_V(V)$: $\nabla_V G_V(V)=\sum_{i=1}^{N}\big\{2x_i^TVu_i^T-(2x^T_i+\pi_i)u_i\}$. 
Subsequently, with the gradient $\nabla_V G_V(V)$, we are able to solve the subproblem (\ref{V}) via the iteration $i=0,I-1$ 
\begin{equation}\label{Proxz1}
	V^{i+1}\leftarrow  \text{ Prox}_{\mu\eta_V\norm{\cdot}_*} \left\{V^i-\eta_V\nabla_V G(V^i)\right\},
\end{equation}
where $\eta_V>0$ is the step size. After the server finished the update for item profile matrix $V$, it can be shared by all users. Then, user $i$ download $V$ for its updating $x_i$ and $u_i$ with order and the problem can be formulated in the following: 
\begin{equation}\label{order}
	\begin{aligned}
		(x_i,u_i)\leftarrow&\underset{x_i,u_i}{\text{argmin }}\norm{x_i-R_{i,:}}^2_2+\lambda\norm{u_i}_2+\frac{\rho}{2}\norm{x_i-u_i\cdot V^T}^2+\langle \pi_i , x_i-u_i\cdot V^T\rangle;
	\end{aligned}
\end{equation}	
Specifically, the update for $x_i$ can be obtained straightforwardly in closed-form. Subsequently,  given current $x_i$, $u_i$ can be updated via standard minimization. Finally, the client $i$ updates $\pi_i$ via: $\pi_i\leftarrow\pi_i+\rho(x_i^T-V\cdot u_i^T)$. 

As is shown, the above ADMM procedure can conveniently solve the MF problem in recommendation system under the federated settings, it should be noted that the steps 2) and 3) can be carried out in parallel on each client. 

\begin{algorithm}[ht]
	\renewcommand{\algorithmicrequire}{\textbf{Input:}}
	\renewcommand{\algorithmicensure}{\textbf{Output:}}
	\caption{3-Block Decentralized ADMM for MF} 
	\label{alg:3bdADMM-MF}
	\begin{algorithmic}[1]
		\STATE \textbf{server input:} The communication round $R$ and iteration number $I$, initial $U$ and $V$, $\alpha$ and $\rho$.
		%\ENSURE $y = x^n$ 
		\STATE \textbf{user $i$'s input:} The local iteration number $T$, initial $\eta$, $x_i$, and $y_{i,j}$. 		
		\FOR{$r=1,\dots,R$}
		\STATE 
		
		{\textbf{Server implements} steps 5-7:}
		\STATE
		Receive  $(2x^T_i+\pi_i,u_i)$ from users $i\in\mathcal{S}$
		\STATE
		Updates $V$ via (\ref{Proxz1}) for $i=0,\cdots,I-1$.
	
		\STATE 
		
		Randomly sample users ${\mathcal{S}}\subseteq[N]$ and inform the users to download $V$.
		
		\STATE
		{\textbf{Users implement} steps 8-11 \textbf{in parallel for} $i\in{\mathcal{S}}$:}
		
		\STATE
		User $i$ download $V$, and performs the update orderly for $(x_i,u_i)$ via (\ref{order}). 
		\STATE updates  $\pi_i$: $\pi_i\leftarrow\pi_i+\rho(x_i^T-V\cdot u_i^T)$
		\STATE User $i$ transmits $(2x^T_i+\pi_i,u_i)$ to the server.

		\ENDFOR
	\end{algorithmic}
\end{algorithm}

	 \section{Optimization}\label{opt}
Federated learning is essentially based on the techniques of distributed optimization  for the machine learning, which is abbreviated as FedOpt in this survey \cite{ DBLP:journals/corr/KonecnyMRR16,DBLP:journals/corr/KonecnyMYRSB16,pmlr-v54-mcmahan17a,10.1145/3298981}. The typical characteristics distinguish FedOpt from classical distributed optimization by the following three aspects:
\begin{packed_itemize}
	\item To begin with,  while FedOpt targets at training a high-quality model from the decentralized data, its primary goal is to ensure the data privacy and security of each participated agent.
	\item Second, the real world applications of FL generally include massive participants, there is high probability that some participant remains inactive, since they maybe in offline, low-battery and power off status, while the typical distributed optimization,  requires all clients synchronously or a large portion of clients  to asynchronously participate in each round update.
	\item Third, due to each client's own fashion, the local data among them are generally statistically heterogeneous, known as non-IID data. In general, it has been shown partial participation and data heterogeneity  bring performance degradation and prevent  learning  from reaching fast convergence in FL scenarios.
\end{packed_itemize}   
As a result, building efficient and effective FedOpt algorithms has recently drawn substantial interests \cite{DBLP:journals/corr/KonecnyMYRSB16,LiXiang2019OtCo,li2018federated,pmlr-v54-mcmahan17a}.

%to  the major areas have been greatly increased, e.g., healthcare \cite{Sheller,HUANG2019103291,BRISIMI201859} and smart city applications \cite{wang2020federated,zheng2021federated}. 

\subsection{Federated Optimization}
In this section, we comprehensively review the existing literatures in FedOpt and characterize FedOpt algorithms from the aspects of search direction accuracy and the alternating direction of multiplier based methods (ADMM).  In particular, the first-order approaches are the most popular choices. \textbf{We extensively review these works and find out their basic paradigms of the local acceleration and the global acceleration.}   With the basic frameworks, we also provide popular techniques, e.g., quantization and sparsification for performance improvement.
%	Federated optimization is essentially a sub-class of the distributed optimization. Hence, in this subsection, we review the distributed  optimization, which may motivate the development for the federated optimization. 
\subsubsection{Federated Problem Formulation}
We first formulate the federated problem. Suppose there are $N$ workers, 
and each worker $i\in[N]$ has the local loss function $f_i(x)$ with its own dataset $\mathcal{D}_i$ containing $n_i$ {samples}. As a result, federated optimization targets at collaboratively solving the following empirical risk minimization problem over $N$ workers and one server:
\begin{equation}\label{fedobj1}
	\underset{x}{\text{min }}f(x) 
	= \frac{1}{N}\sum\nolimits_{i=1}^{N}f_i(x),
\end{equation}
where  $f$ is the averaged loss function. The model as the optimization variable satisfies $x\in\mathbb{R}^d$. Moreover, the above functions satisfy $f:\mathbb{R}^d\rightarrow\mathbb{R}$, $f_i:\mathbb{R}^d\rightarrow\mathbb{R}$.

\begin{figure*} 
	\begin{minipage}[t]{0.9\linewidth} % 如果一行放2个图，用0.5，如果3个图，用0.33
		\centering
		\begin{subfigure}{1\textwidth}
			\centering
			\includegraphics[width=1\linewidth]{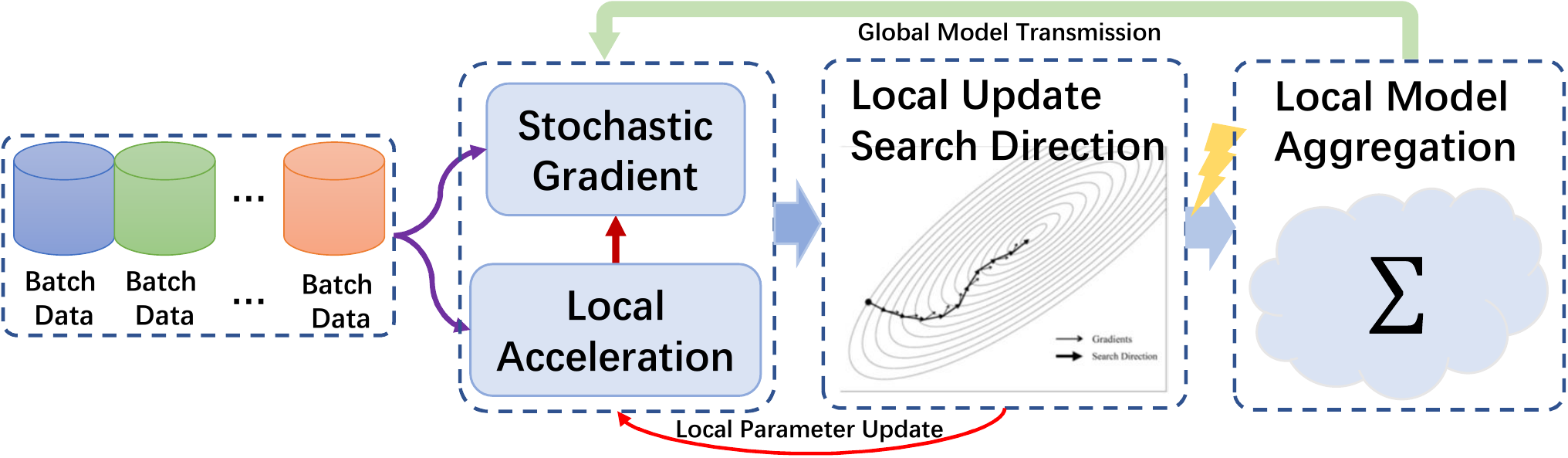}
		\end{subfigure}\\
		\subcaption{Local acceleration scheme.}
	\end{minipage} \hfill	
	\begin{minipage}[t]{0.9\linewidth} % 如果一行放2个图，用0.5，如果3个图，用0.33
		\centering
		\begin{subfigure}{1\textwidth}
			\centering
			\includegraphics[width=1\linewidth]{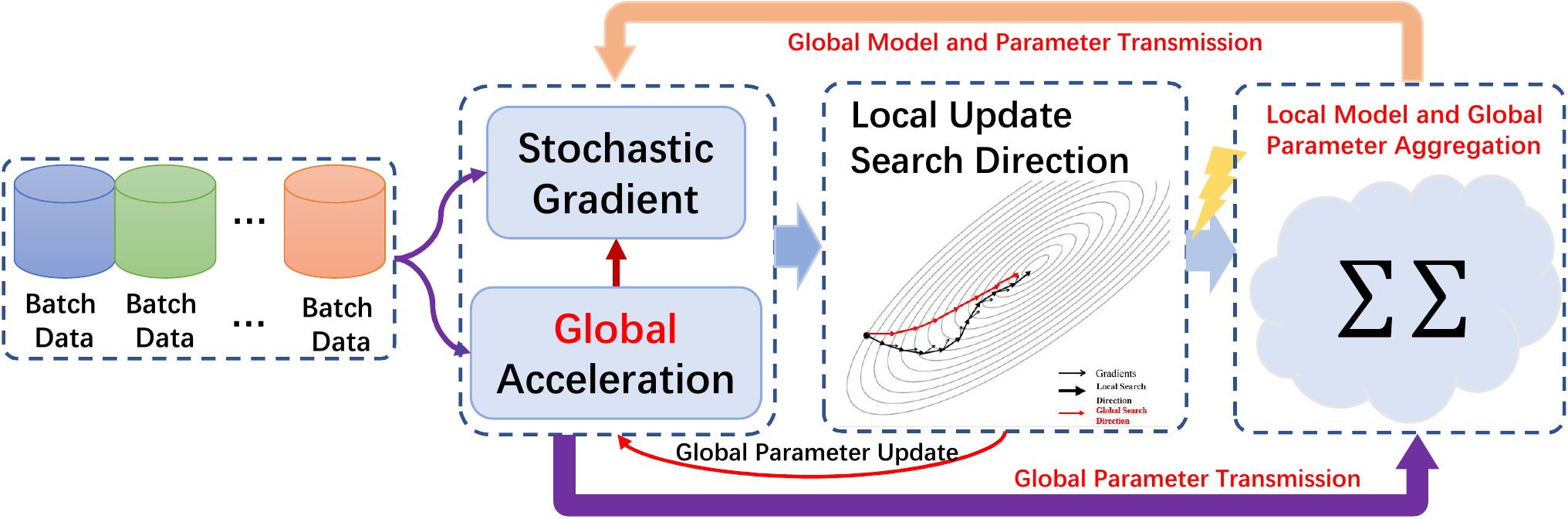}
		\end{subfigure}\\%\\
		\subcaption{Global acceleration scheme.}
	\end{minipage} 	
	\caption{ Illustration of local and global updates in FL.  (a). First, the local dataset block is randomly chosen for the gradient calculation; second, obtain the search direction by combining the local acceleration, which is subsequently used for local model update; third, the local acceleration is updated using local information; fifth, the local model is transmitted to the server for global aggregation. (b). Different with the local acceleration scheme, the global acceleration scheme also transmits the global acceleration for global aggregation.}
	\label{fig:localglobalacceleration}
\end{figure*}
\para{Implementation Strategy.}
In terms of the FedOpt implementation, there are two types: synchronous and asynchronous algorithms. In synchronous distributed optimization, all the workers perform the local updates and then transmit their corresponding information (e.g., gradients and variables, etc..) to  the central server, which subsequently performs aggregation \cite{10.5555/2627435.2638571,NIPS2010_abea47ba,10.5555/1857999.1858068,NIPS2009_d81f9c1b}. However, due to the device heterogeneity in workers, the server always needs to wait  for the slowest worker to push its information. This has significantly limited the scalability. On the contrary, in asynchronous algorithms, the servers conduct the updates upon  receiving the information from the specified bunch of workers with the fastest speed, while it generally does not wait for the slower workers (or seldom wait in some scenarios \cite{pmlr-v32-zhange14}).

\subsubsection{First-order Scenario}\label{fos}
In this scenario, the first-order information is adopted for local search direction. Specifically, momentum and variance reduction techniques can be used for acceleration. According to the acceleration position, we divide them into local and global acceleration schemes respectively. We also provide the instantiation of these two scenarios in the supplementary material.

\para{Local acceleration.} While the acceleration techniques such as the momentum and variance reduction have been widely utilized to accelerate SGD, e.g., Adam, AdaGrad and SVRG  etc., they are also employed in the  distributed SGD to speed up the convergence. Specifically, mimicking the adaptation of SGD to FedAvg, the momentum gradient descent method which is a variant of SGD has been extended to the local update step, then the global performs aggregating all the local models and the momentum terms,  where the resultant algorithm is called MFL \cite{LiuWei2020AFLv}.   {Similarly, FedAdam \cite{FedAdam} extend Adam \cite{Adam} to FedOpt.}  However, these methods still suffer from the {model divergence} problem \cite{WangShiqiang2019AFLi,ZhaoYue2018FLwN,LiXiang2019OtCo,DBLP:conf/aaai/YuYZ19}.	This can be due to the reason of the multiple local update steps in FedOpt which will make the local model fast approach to the local optimum of  the local loss function instead of the global one \cite{SCAFFOLD}. Moreover, since only a small portion of the clients participate in the update on each communication round, the aggregated model will be further biased from the optimal global one. Intuitively, it can be attributed to the global aggregation  with  only   the partial information from the participated clients.  These algorithms share the common patterns - local acceleration strategy, which can be merged together in the Algorithm \ref{alg:localacceleration}. 
\begin{algorithm}[ht]
	\renewcommand{\algorithmicrequire}{\textbf{Input:}}
	\renewcommand{\algorithmicensure}{\textbf{Output:}}
	\caption{Local Acceleration Framework} 
	\label{alg:localacceleration}
	\begin{algorithmic}[1]
		\STATE \textbf{server input:} The communication round $R$, initial global model $x$.
		%\ENSURE $y = x^n$ 
		\STATE \textbf{client $i$'s input:} The local iteration number $T$, local model $x_i$, and the acceleration  $\mathcal{L}_i$. 		
		\FOR{$r=1,\dots,R$}
		\STATE 
		
		{\textbf{Server implements} steps 5-6:}
		
		\STATE
		Updates the global model $x$ by aggregating $x_i$ for $i=1,\dots,N$.   		
		
		\STATE 
		
		Sample clients ${\mathcal{S}}\subseteq[N]$ and	transmit  $x$ to each client $i\in{\mathcal{S}}$.
		
		\STATE
		{\textbf{Clients implement} steps 8-15 \textbf{in parallel for} $i\in{\mathcal{S}}$:}
		
		\STATE
		After receiving, initializing the local model $x_i$ with $x$.
		
		\FOR{$t=1,\dots,T$}
		%		\IF{$0\in\mathcal{I}^{s+1}_{t+1}$} % If 语句，需要和EndIf对应
		
		\STATE Sample a data sample from local dataset and calculate the stochastic gradient $g_i$.
		\STATE
		Obtain the search direction with $(g_i, \mathcal{L}_i)$: $\mathcal{V}_i(g_i, \mathcal{L}_i)$.

		\STATE
		Local model update: $x_i\leftarrow x_i-\eta \mathcal{V}_i(g_i, \mathcal{L}_i)$.
		\STATE
		The acceleration update via $\mathcal{U}_i$:  $\mathcal{L}_i\leftarrow \mathcal{U}_i(x_i, g_i, \mathcal{L}_i)$.  
		
		\ENDFOR    		
		\STATE Client $i$ transmits the local model to the server.

		\ENDFOR
	\end{algorithmic}
\end{algorithm}

\para{Global acceleration.} For improving the performance,  the control variate for variance reduction strategy has been adopted, i.e., VRL-SGD, which has shown faster speed even with non-IID data \cite{VRLSGD}. However, VRL-SGD does not support the client sampling which is more practical in FL. Furthermore, with the control variate to include as more information from clients as possible, SCAFFOLD has shown the significant performance improvement for the data heterogeneity problem in FedOpt \cite{SCAFFOLD}. Its core notion is to estimate the full gradient for the local search direction.
Most recently, a framework called Mime is proposed \cite{MIME}, which adapts popular centralized algorithms (e.g., SGD, Adam etc.) to FedOpt.  

We summarize these algorithms into the general global acceleration strategy in Algorithm \ref{alg:globalacc}  to provide the motivation for the development of novel algorithms. It can be seen these algorithms are different with the local acceleration scheme in that the acceleration adopts the global information. We further illustrate the global acceleration scheme in Fig. \ref{fig:localglobalacceleration}.

\begin{algorithm}[ht]
	\renewcommand{\algorithmicrequire}{\textbf{Input:}}
	\renewcommand{\algorithmicensure}{\textbf{Output:}}
	\caption{Global Acceleration Framework} 
	\label{alg:globalacc}
	\begin{algorithmic}[1]
		\STATE \textbf{server input:} The communication round $R$, initial global model $x$ and the acceleration $\mathcal{M}$.
		%\ENSURE $y = x^n$ 
		\STATE \textbf{client $i$'s input:} The local iteration number $T$, local model $x_i$, and the acceleration  $\mathcal{M}_i$. 		
		\FOR{$r=1,\dots,R$}
		\STATE 
		
		{\textbf{Server implements} steps 5-6:}
		
		\STATE
		Updates   $x$ and  $\mathcal{M}$ by aggregating $x_i$ and $\mathcal{M}_i$ respectively for $i=1,\dots,N$.   		
		
		\STATE 
		
		Sample clients ${\mathcal{C}}\subseteq[N]$ and	transmit  $x$  and  $\mathcal{M}$ to each client $i\in{\mathcal{C}}$.
		
		\STATE
		{\textbf{Clients implement} steps 8-15 \textbf{in parallel for} $i\in{\mathcal{C}}$:}
		
		\STATE
		After receiving, initializing the local model $x_i$ with $x$ and $\mathcal{M}_i$ with $\mathcal{M}$.
		
		\FOR{$t=1,\dots,T$}
		%		\IF{$0\in\mathcal{I}^{s+1}_{t+1}$} % If 语句，需要和EndIf对应
		
		\STATE Sample a data sample from local dataset and calculate the stochastic gradient $g_i$.
		\STATE
		Obtain the search direction with $(g_i, \mathcal{M}_i)$: $\mathcal{V}_i(g_i, \mathcal{M}_i)$.

		\STATE
		Local model update: $x_i\leftarrow x_i-\eta \mathcal{V}_i(g_i, \mathcal{M}_i)$.
		\STATE
		The acceleration update via $\mathcal{U}_i$:  $\mathcal{M}_i\leftarrow \mathcal{U}_i(x_i, g_i, \mathcal{M}_i)$.  
		
		\ENDFOR    		
		\STATE Client $i$ transmits $(x_i,\mathcal{M}_i)$ to the server.

		\ENDFOR
	\end{algorithmic}
\end{algorithm}

\para{Quantization} Despite the numerous advantages of distributed computing via FedOpt, significant limitations exist in the communication process, since massive workers transmit the gradients or the models for aggregation in the server, resulting in communication overhead. Quantization technique is a straightforward idea for compressing the communication quantities. In \cite{pmlr-v108-reisizadeh20a}, 
a quantized version of FedAvg, known as FedPaq, is proposed for reducing the message overload. Another strategy, called QSGD \cite{DBLP:conf/nips/AlistarhG0TV17},   quantizes the gradient in each  local worker with a specified random quantization operator and it costs only $(2+\log(2N+1))Nd$ in each round communication.  Jeremy et al. \cite{DBLP:conf/icml/BernsteinWAA18} proposes SIGNSGD and finds  only transmitting the sign of the gradient on each worker can still get the
SGD-level convergence rate, while it only costs $2Nd$ bits in each round communication.  As for the acceleration strategy for QSGD and SIGNSGD,   SVRG is applied in QSGD,  and the momentum method is adopted to SIGNSGD, the resultant algorithms are QSVRG and  SIGNUM respectively.  However, as for the QSGD, its convergence is based
on unrealistic assumptions and can diverge in practice. To remedy this issue, 
Nesterov’s momentum has been incorporated to the general distributed SGD with two-way compressions on both the workers and server \cite{NEURIPS2019_80c0e8c4}, which has demonstrated numerically reducing the communication by about $32\times$ and can converge as quickly as full-precision distributed momentum SGD to reach the same testing accuracy.

\para{Sparsification} While the  quantization is effective in reducing the communication overhead, the sparsification technique is another choice for the communication compression, where only the most important and information-preserving gradient entries are sent when using gradient sparsification technique. 

Specifically, Dan et al. \cite{NEURIPS2018_31445061} and Stich et al. \cite{NEURIPS2018_b440509a} study sparsifying gradients by magnitude with local error correction and selecting top $K$ components.  Another  algorithm only transmits the large
entries to the server \cite{onebit}, which shows the significant reduction in the communication overhead \cite{DBLP:journals/corr/AjiH17}.  While the sparsification can bring performance degradation, deep gradient compression \cite{DBLP:journals/corr/abs-1712-01887} combines it with other techniques including momentum correction, local gradient clipping, and momentum factor masking to achieve higher performance with considerable communication cost reduction.  In fact, these strategies can deliver good results, but they have a few limitations. First, they currently have no guarantees of convergence; second, although the workers can perform the compression to high rate, when the server implements aggregation of the disjoint top $K$, it will again become dense.    To alleviate the problem, Sketched-SGD has been proposed by introducing the sketching technique in distributed SGD, which is motivated by the success of sketching methods in sub-linear/streaming algorithms \cite{NEURIPS2019_75da5036}. The main idea of Sketched-SGD is  communicating sketches with top $K$ sparsification on both the server and workers. For the communication overhead reduction, Sketched-SGD only communicates $O(logd)$ information, while the general gradient compression methods transmits	$O(d)$. 

Although sparsifying the transmitted gradients can improve the communication efficiency, it also increases the gradient variance, which may slow down the convergence speed.  To mitigate the problem, Wangni et al. \cite{NEURIPS2018_3328bdf9} proposes a gradient sparsification technique to  find the optimal tradeoff between sparsity and the gradient variance. Specifically, the key idea is to randomly drop out each coordinate $g^j_i$ of the stochastic gradient vector $g_i$ by probability $1-p_j$ on the $i$th worker, and appropriately amplify the remaining coordinates to ensure that the sparsified gradient is unbiased. Then the variance of the quantized gradient $Q(g_i)$ can be bounded by $\mathbb{E}\sum_{j=1}^{d}[Q(g^j_i)^2]=\sum_{j=1}^{d}\nicefrac{(g^j_i)^2}{p_j}.$
In addition, the expected sparsity $\mathbb{E}\norm{Q(g_i)}_0$ can be computed by $\mathbb{E}\norm{Q(g_i)}_0=\sum_{j=1}^{d}p_j$. Subsequently, the target is to obtain the gradient as sparsified as possible with the variance as small as possible, which can be formulated as the following problem: $\underset{p}{\text{min}} \sum_{j=1}^{d}p_j, \text{ s.t. } \sum_{j=1}^{d}\nicefrac{(g^j_i)^2}{p_j}\leq \epsilon.$
\tiny
\begin{table*}[ht]
	%  \resizebox{1\textwidth}{!}{%
		\begin{center}
			\begin{tabular}{l|l|l|l|l|l|l|l|l|l}
				\hline
				\textbf{Algorithm}  &\textbf{\tabincell{l}{Local\\ update }}  &\textbf{\tabincell{l}{Local\\steps }}  &\textbf{\tabincell{l}{Local\\ complexity }}  &\textbf{\tabincell{l}{Transmission\\ quantity }}  &\textbf{\tabincell{l}{Global\\ update }}   &\textbf{\tabincell{l}{Convergence\\ speed }}     &\textbf{\tabincell{l}{Dealing \\with M.D. }} &\textbf{SC/AS} &\textbf{PCP} \\ \hline
				%\multirow{10}{*}{Synrony} 
				
				VRL-SGD   &\tabincell{l}{Variance\\ reduced\\ SGD}  &\tabincell{l}{Multiple}  &\tabincell{l}{$\mathcal{O}(T\cdot d)$}  &\tabincell{l}{Local and\\ Global models
				}   &\tabincell{l}{Weight\\ averaging }     &\tabincell{l}{$\mathcal{O}\big(\frac{1}{R}\big)$}   &\tabincell{l}{N/A} &SC & \tabincell{l}{$\times$}    \\ \cline{1-10}

				MFL  &MGD  &Multiple  &$\mathcal{O}(Td)$  &\tabincell{l}{Local model\\ and momentum  \\Global model \\and momentum }   &\tabincell{l}{Weight\\ averaging }   &$\mathcal{O}(\frac{1}{R})$    &N/A &SC &$\times$ \\ \cline{1-10} 
				
				QSGD  &\tabincell{l}{Calculate   \\ quantized\\ gradient}  &Single  &$\mathcal{O}(d)$   &\tabincell{l}{Quantized   \\gradient}    &\tabincell{l}{Global\\SGD}    & $\mathcal{O}(\frac{1}{R})$   &N/A &SC &$\times$  \\ \cline{1-10} 
				
				FedAvg  &SGD  &Multiple  &\tabincell{l}{$\mathcal{O}(T\cdot d)$}  &\tabincell{l}{Local model,\\
					Global model}   &\tabincell{l}{Weight\\ averaging }      &\tabincell{l}{$\mathcal{O}\big(\frac{1}{R\cdot T}\big)$}   &\tabincell{l}{$\checkmark$}    &AS & \tabincell{l}{$\times$} \\ \cline{1-10}
				
				SCAFFOLD  &\tabincell{l}{Variance\\ reduced\\ SGD}   &Multiple  &\tabincell{l}{$\mathcal{O}(T\cdot d)$}   &\tabincell{l}{Local model\\ and gradient,\\
					Global model \\and gradient
				}  &\tabincell{l}{Weight\\ averaging }   &\tabincell{l}{$\mathcal{O}\big(\frac{1}{R\cdot T}\big)$}    &\tabincell{l}{$\checkmark$} &AS &\tabincell{l}{$\checkmark$} \\ \cline{1-10} 
				
				FedProx  &PGD  &Single &$\mathcal{O}(d)$   &\tabincell{l}{Local model,\\
					Global model}     &\tabincell{l}{Weight\\ averaging }   &$\mathcal{O}\big(\frac{1}{R}\big)$   &limited &AS &$\checkmark$  \\ \cline{1-10} 
				
				DIANA  &\tabincell{l}{Calculate   \\ quantized\\ gradient\\difference}   &Single  &$\mathcal{O}(d)$   &\tabincell{l}{Quantized \\ gradient  \\difference, \\Global model\\and gradient\\ shift}   &\tabincell{l}{Global\\ PGD and\\gradient\\shift\\aggregation}      &\tabincell{l}{$\mathcal{O}{(\frac{1}{R})}$  and \\linear\\ convergence \\rate}    &N/A &SC &$\times$ \\ \cline{1-10}

				MimeSVRG  & \tabincell{l}{Variance\\ reduced\\ SGD}  &Multiple  &$\mathcal{O}(2Td)$  &\tabincell{l}{Local model\\and gradient,\\
					Global model \\and aggregated\\ gradient
				}  &\tabincell{l}{Weight\\ averaging }    &\tabincell{l}{$\mathcal{O}\big(\frac{1}{R\cdot T}\big)$}     &$\checkmark$&AS &$\checkmark$  \\ \cline{1-10} 
				
			\end{tabular}
		\end{center}
		%}
	\caption{Representative federated optimization methods based on first-order gradient.  Specifically, M.D. is acronym for model divergence, synchony and asynchrony are abbreviated as SC and AS respectively, the abbreviation of partial client participation is PCP.}
	\label{Tab:opt1}
\end{table*}
\normalsize

It should be emphasized that none of the above compression (i.e., quantization and sparsification) based methods can learn the gradients, preventing them from converging to the true optimum in batch mode, making them incompatible with non-smooth regularizers, and slowing their convergence.  To mitigate the problem, DIANA is proposed by compressing the gradient difference and performing the proximal minimization for dealing with the non-smooth regularizers in the local \cite{DBLP:journals/corr/abs-1901-09269}.  Another strategy ADIANA combines the acceleration/momentum and compression strategies \cite{pmlr-v119-li20g}.

\subsubsection{Second-order Scenarios} The second-order scheme has attracted attentions due to the higher accuracy in the search direction. Although   they generally need more computation per iteration, they 
require fewer iterations to achieve similar results \cite{doi:10.1137/16M1080173}.   

DANE  minimizes the linear approximation of the objective function with Bregmen divergence measurement \cite{pmlr-v32-shamir14}. It has been found that each computational machine in DANE implicitly uses its local Hessian in quadratic objective cases, while no Hessians are explicitly computed. Wang et al. \cite{NEURIPS2018_dabd8d2c} proposes a distributed  Newton-type    optimization method GIANT. Technically, GIANT realizes the local approximated Newton directions  by the conjugate gradient descent method (CG), which involves only Hessian-vector products in CG iterations.  In \cite{NEURIPS2019_9718db12}, a novel method based on   second-order information is proposed, known as DINGO and is derived
by optimization of the gradient’s norm as a surrogate function for the objective function.     For further supporting second-order methods towards FL,  Safaryan et al. \cite{DBLP:conf/icml/SafaryanIQR22} proposes FedNL framework, and its main ingredient is to   reuse past Hessian information and build the next Hessian estimate by updating the current estimate. Moreover, for reducing communication overload, the compression technique is adopted for Hessian compression.  Furthermore, Agafonov et al. \cite{NEURIPS2022_000c076c} proposes the similar strategy FLECS-CGD, which also compresses the gradient information. However, they both suffer from high computation complexity, making them unpractical in mobile applications with low power consumption. To mitigate this issue, Ma et al. \cite{DBLP:journals/corr/abs-2206-09576} proposes FedSSO approach, it updates the local model via SGD, and pushes the Hessian update and global model update to the server.

\subsubsection{ADMM} Another class of FedOpt algorithm is based on the alternating direction method of multipliers (ADMM) \cite{DBLP:journals/ftml/BoydPCPE11}, which is simple and efficient.   ADMM  decouples the complicated problems with the coupling constraint(s) into smaller subproblems coordinated by a master problem. Moreover, it utilizes the augmented Lagrangian function \cite{DBLP:journals/mp/Powell78,Hestenes}, which penalizes the constraint equality with a quadratic term, making it advantageous in numerical stability.  Therefore, ADMM is adept at solving the complicated problems with high degree of parallelization, and exhibits linear scaling as data is processed in parallel across devices, making it a natural fit in the large-scale distributive applications.   In recent years, ADMM has attracted a considerable amount of attention in both practical and theoretical aspects \cite{DBLP:conf/icml/TaylorBXSPG16,DBLP:journals/ftml/BoydPCPE11}.    With simple and straightforward modification, ADMM can be adapted to FedOpt.	

\tiny
\begin{table*}[ht]
	%  \resizebox{1\textwidth}{!}{%
		\begin{center}
			\begin{tabular}{l|l|l|l|l|l|l|l|l|l}
				\hline
				\textbf{Algorithm}  &\textbf{\tabincell{l}{Local\\ update }}  &\textbf{\tabincell{l}{Local\\steps }}  &\textbf{\tabincell{l}{Local\\ complexity }}  &\textbf{\tabincell{l}{Transmission\\ quantity }}  &\textbf{\tabincell{l}{Global\\ update }}   &\textbf{\tabincell{l}{Convergence\\ speed }}     &\textbf{\tabincell{l}{Dealing \\with M.D. }} &\textbf{SC/AS} &\textbf{PCP}  \\ \hline
				%\multirow{10}{*}{Synrony} 

				FedNL  &\tabincell{l}{Calculate  \\ compressed \\
					Hessian \\
					and gradient}  &Single  &$\mathcal{O}(d^2)$  & \tabincell{l}{Compressed \\Hessian and\\the gradient}.
				&\tabincell{l}{Global model\\  and Hessian \\update}  &\tabincell{l}{sub-linear\\ convergence.}    &N/A &AS &$\checkmark$  \\ \cline{1-10}
				
				FedSSO  &\tabincell{l}{SGD}  &multiple  &$\mathcal{O}(Td)$  & \tabincell{l}{Local and \\global models}.
				&\tabincell{l}{BFGS Hessian\\ and global \\model update}  &\tabincell{l}{$\mathcal{O}(1/R)$}    &N/A &AS &$\checkmark$  \\ \cline{1-10}
				
				FLECS-CGD  &\tabincell{l}{Calculate  \\ compressed \\
					Hessian \\
					and gradient}  &Single  &$\mathcal{O}(d^2)$  & \tabincell{l}{Compressed \\Hessian and\\the gradient\\difference}.
				&\tabincell{l}{Global model\\  and Hessian \\update}  &\tabincell{l}{sub-linear\\ convergence.}    &N/A &AS &$\checkmark$ 
				
				\\ \hline
			\end{tabular}
		\end{center}
		%}
	\caption{Representative federated optimization methods based on second-order gradient.}
	\label{Tab:opt2}
\end{table*}	
\normalsize

For algorithmic development, Goldstein et al. \cite{DBLP:journals/siamis/GoldsteinOSB14} has incorporated Nesterov's acceleration strategy \cite{Nesterov1983AMF} in ADMM for solving the separate subproblem.  It has reached the convergence speed of $\mathcal{O}(\nicefrac{1}{R^2})$ in strongly convex problems. With simple modification, the accelerated ADMM can fit in FedOpt. Another strategy is based on Anderson acceleration \cite{10.1145/3355089.3356491}, it targets at speeding up the convergence of a fixed-point iteration \cite{doi:10.1137/10078356X} via the quasi-Newton method by adopting previous iterates by approximating its	inverse Jacobian \cite{EYERT1996271,https://doi.org/10.1002/nla.617}. It has shown its superior performance in the nonconvex problems of computer graphics tasks.  Considering the topological network, DADMM introduces the undirected graph for modeling  the network topology of agents, and the decentralized learning is performed over the agents \cite{DBLP:journals/tsp/ShiLYWY14}. For further improving the computational efficiency in DADMM, Ling et al. \cite{DBLP:journals/tsp/LingSWR15} has proposed a linearized version of DADMM (DLM). Both DADMM and DLM have been demonstrated the linear convergence rate under the  assumption of strong convexity in the objective function.  For supporting ADMM in FL, Zhou et al. \cite{DBLP:journals/corr/abs-2110-15318} proposes CEADMM approach, its key idea is to modify the classical ADMM by update the local model and Lagrangian multiplier multiple times before transmitting to the server. 
\tiny
\begin{table*}[ht]
	%  \resizebox{1\textwidth}{!}{%
		\begin{center}
			\begin{tabular}{l|l|l|l|l|l|l|l|l|l}
				\hline
				\textbf{Algorithm}  &\textbf{\tabincell{l}{Local\\ update }}  &\textbf{\tabincell{l}{Local\\steps }}  &\textbf{\tabincell{l}{Local\\ complexity }}  &\textbf{\tabincell{l}{Transmission\\ quantity }}  &\textbf{\tabincell{l}{Global\\ update }}   &\textbf{\tabincell{l}{Convergence\\ speed }}     &\textbf{\tabincell{l}{Dealing \\with M.D. }} &\textbf{SC/AS} &\textbf{PCP} \\ \hline
				%\multirow{10}{*}{Synrony} 
				
				CEADMM  &\tabincell{l}{Primal \\and  dual\\ updates}  &Multiple  &$\mathcal{O}(Td)$  &\tabincell{l}{Local model\\ and Lagrangian \\multiplier, \\global model}  &\tabincell{l}{Primal\\update}   &N/A   &N/A &AS & $\checkmark$  \\ \cline{1-10} 
				
				Fast ADMM  &\tabincell{l}{Primal \\and  dual\\ updates with \\acceleration}  &Single  &$\mathcal{O}(d)$  &\tabincell{l}{Local model\\ and gradient, \\ Lagrangian \\multiplier, \\global model\\ and gradient}  &\tabincell{l}{Primal\\update}   &$\mathcal{O}(\frac{1}{R^2})$   &N/A &SC & $\times$  \\ \cline{1-10} 
				
				AsyncADMM  &\tabincell{l}{Primal \\and  dual\\ updates with \\acceleration}    &Single  &$\mathcal{O}(d)$  &\tabincell{l}{Local model\\and gradient,\\
					Global model \\and gradient}    &\tabincell{l}{Primal\\update}   &$\mathcal{O}\big(\frac{1}{R}\big)$    &N/A  &AS &$\times$ \\ \hline
				
			\end{tabular}
		\end{center}
		%}
	\caption{Representative federated optimization methods based on ADMM. }
	\label{Tab:opt12}
\end{table*}
\normalsize
%\subsubsection{Future direction} Recently, ADMM has shown its extraordinary flexibility and convenience for solving a class of problems in the graph networks of multiple agents with general graph topology \cite{https://doi.org/10.48550/arxiv.2010.10421,https://doi.org/10.48550/arxiv.2210.17241}. Compared with the traditional structure of FL, namely multiple clients coordinated by a single server, the optimization over a graph network can be more practical in real applications, e.g., by cooperative control of connected and automated vehicles via distributed optimization, it has shown saving up to 32.53\% fuel consumption \cite{https://doi.org/10.48550/arxiv.2210.13171}. We believe by developing suitable distributed optimization algorithms over a graph network, FL will promote the significant development in various areas.  

\section{Privacy and Security}\label{pas}

\begin{figure}
	\centering
	\includegraphics[width=1\linewidth]{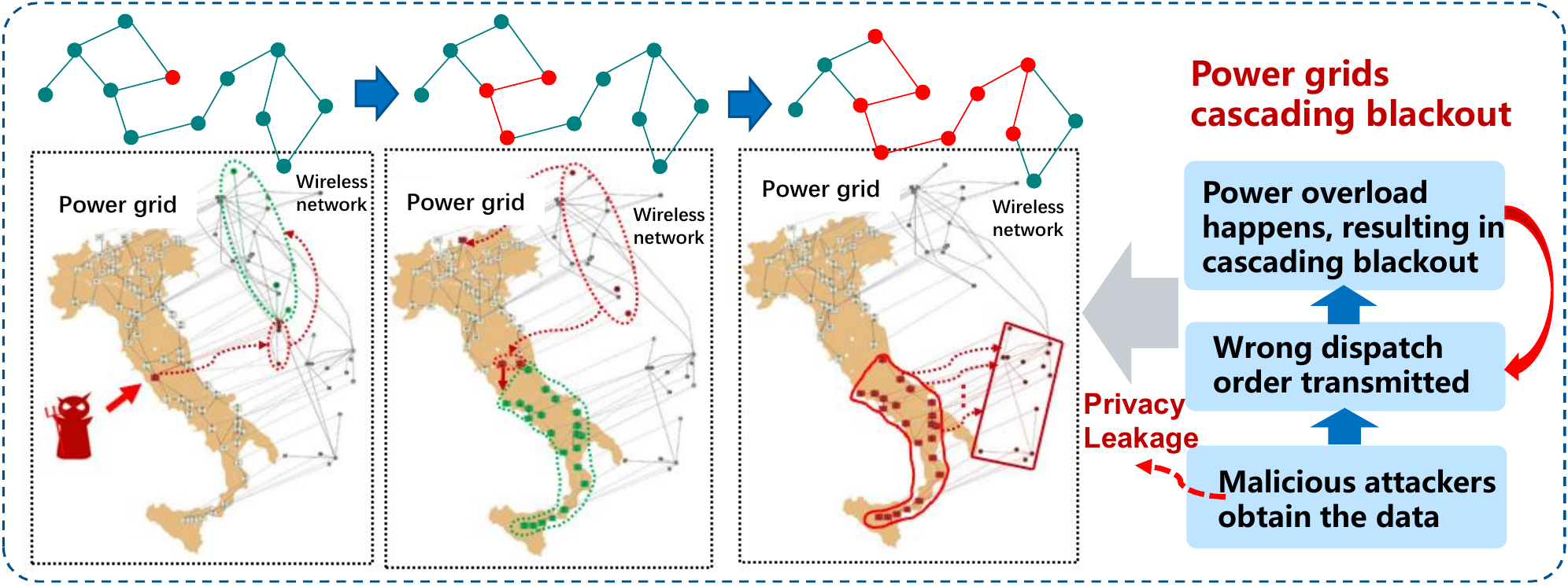}
	\caption{Cascading blackout resulted by information leakage, directly impacting national security.}
	\label{fig:blackout}
\end{figure}

The privacy and security issues are important for the FL system, without proper protection, they may result in the consequence of huge losses, e.g., malicious attack in power grid  results in catastrophic blackout shown in Fig. \ref{fig:blackout}.   In general, privacy and security in FL can be divided into two categories, i.e., data privacy and model security. Specifically, in terms of the privacy issues, existing studies mainly focus on modeling the threat model of leaking information about the data of the participants in the process of FL and developing privacy-preserving mechanisms to prevent privacy leakage. On the other hand, in terms of the security issues, existing studies mainly focus on modeling the treat model of influencing the performance of the model learned based on FL against malicious participants. We will introduce each problem in terms of their corresponding attack models and defence methods, respectively.

\para{Attack methods.} The goal of the attackers in terms of privacy issues is to reveal the information about the data of the participants in the process of FL. Then, the attack methods can be divided into different categories based on what information is leaked. Here, we list major categories of attack methods as follows:
\begin{packed_itemize}
\item \textbf{Reconstruction Attack:} The goal of this kind of attack is to reconstruct the data samples of the participants based on the transmitted parameters observed by the attacker.
\item \textbf{Membership Inference Attack:}  Different from the reconstruction attack, the goal of this kind of attack is to infer whether a given data sample was including by the dataset of a particular participant.
\end{packed_itemize}

\iffalse
\begin{mdefinition}[Membership Inference Attack]\label{def:miattack}
A randomized mechanism $\mathcal{M}: \mathcal{D} \rightarrow \mathcal{O}$ satisfies $(\epsilon,\delta)$-differential privacy if and only if, for arbitrary adjacent datasets $D_1$ and $D_2$, and subset $O\in\mathcal{O}$, we have $Pr(\mathcal{M}(D_1)\in O)\leq e^\epsilon Pr(\mathcal{M}(D_2)\in O)+\delta$.
\end{mdefinition}
\fi

A number of existing studies focus on designing different attack methods to reveal the privacy leakage risks of FL systems~\cite{nasr2019comprehensive, geiping2020inverting}.
Nasr et al.~\cite{nasr2019comprehensive} conduct a comprehensive analysis of risk of data privacy leakage in terms of the membership inference attack in both centralized learning and FL systems.
Geiping et al.~\cite{geiping2020inverting} focus on the reconstruction attack, and they show that multiple separate input images can be reconstructed from their average gradient in practice, where they propose an privacy attack strategy based on a magnitude-invariant loss along with optimization strategies, indicating that the FL algorithms do not have innate privacy-preserving property and the security guarantee is sill necessary, e.g., the provable differential privacy.
\subsection{Data Privacy}
\para{Defence method.} 
The mainstream defence methods against privacy attacks are based on perturbation mechanisms, random hash mechanisms, homomorphic encryption (HE), and security multi-party computation (SMC), etc.

Specifically, the perturbation mechanisms protect the privacy of participants by adding random noise to their data or released parameters.
Random hash mechanisms map the input data into random features while keep some properties of the input data unchanged. 
These two kinds of mechanisms all fall into random mechanisms. In order to provide a criterion to measure the privacy protection level, these random mechanisms are usually required to satisfy the differential privacy, which is defined as follows.

\begin{mdefinition}[$(\epsilon,\delta)$-differential privacy]\label{def:a2}
A randomized mechanism $\mathcal{M}: \mathcal{D} \rightarrow \mathcal{O}$ satisfies $(\epsilon,\delta)$-differential privacy if and only if, for arbitrary adjacent datasets $D_1$ and $D_2$, and subset $O\in\mathcal{O}$, we have $Pr(\mathcal{M}(D_1)\in O)\leq e^\epsilon Pr(\mathcal{M}(D_2)\in O)+\delta$.
\end{mdefinition}

As we can observe from the definition, $(\epsilon,\delta)$-differential privacy requires a certain level of indistinguishability of the input based on the observed output of the randomized mechanism, i.e., the ratio of the likelihood of generating the same output by the two inputs is bound by $e^\epsilon$ with at most the probability of $\delta$ to not satisfy this constraint. Thus, the parameters $(\epsilon,\delta)$ quantify the hardness of reconstructing the input based on the output of the randomized mechanism.

On the other hand, homomorphic encryption is a technique that allows us to directly implement computing on the ciphertext and guarantee the correctness of the obtain results after decryption.
On the other hand, security multi-party computation is another technique that allows us to obtain the correct computation results without leaking individual private data.
Based on these methods, we can also preserve user privacy in the computing process of FL.

Sarpatwar et al.~\cite{sarpatwar2019differentially} protect the privacy of participants by learning their data summarization distributedly with differential privacy guarantee, which is solved based on two hash functions based on the Rahimi-Recht Fourier features~\cite{rubinstein2012learning} and MWEM method~\cite{hardt2012simple}.
Jayaraman et al.~\cite{jayaraman2018distributed} propose two perturbation mechanisms, i.e., the output perturbation method and the gradient perturbation method, of distributed learning to achieve differential privacy, where are all combined with the secure multi-party computation (SMC). Specifically, in the output perturbation method, local models of different participants are aggregated once, and the differential privacy noise is added to the aggregated model before it is revealed. As for the gradient perturbation method, local gradients of different participants are aggregated iteratively to collaboratively train a model, and similarly the differential privacy noise is added to the aggregated gradients before it is revealed.
Dubey et al.~\cite{dubey2020differentially} investigate the FL problem of solving a contextual linear bandit based on cooperation of multiple participants, whose communication privacy should be protected, and they solve this problem  based on upper confidence bounds (UCB)~\cite{abbasi2011improved, li2010contextual} and differential private perturbations ~\cite{shariff2018differentially}.
Sun et al.~\cite{sun2021defending} focus on defending the privacy leakage in the vertical FL by proposing a framework containing three modules of adversarial reconstruction, noise
regularization, and distance correlation minimization. 

\tiny
\begin{table*}[t]
	%  \resizebox{1\textwidth}{!}{%
		\begin{center}
			\begin{tabular}{l|l|l|l|l}
				\toprule
				\textbf{Ref.} & \textbf{Privacy/Security} & \textbf{FL Tasks} &  \textbf{Categories}  &\textbf{Idea}   \\
				\midrule
				
				\cite{geiping2020inverting} & Privacy & Information disclosure  & Attack Method &  Magnitude-invariant loss.\\ \cline{1-5} %Inverting Gradients~
				\cite{sarpatwar2019differentially}        & Privacy & Data Summarization & Defence Method  & Random Hash  \\ \cline{1-5}  %Hash-based~
				\cite{jayaraman2018distributed}          & Privacy & Model optimization & Defence Method & SMC, Add perturbation  \\ \cline{1-5} %Perturbation\
				\cite{dubey2020differentially} & Privacy & Federated contextual bandit & Defence Method & Upper confidence  bounds (UCB), differential  private perturbations  \\ \cline{1-5}
				\cite{sun2021defending} & Privacy & Model optimization & Defence Method & Adversarial reconstruction, noise regularization, distance correlation minimization\\ \cline{1-5}
				\cite{bhagoji2019analyzing} & Security & Model optimization & Attack Method & Optimization based on indicators  of  stealth \\ \cline{1-5} %Stealth Model Poisoning 
				\cite{blanchard2017machine} & Security & Model optimization  &  Defence Method &  Distance-based aggregation rule   \\ \cline{1-5}
				\cite{yin2018byzantine} & Security & Model optimization  & Defence Method & Media and trimmed mean operations of the gradient \\ \cline{1-5}
				\cite{li2020byzantine} & Security & Multi-task learning & Defence Method   &  Accumulated loss  \\ \cline{1-5}
				\cite{ghosh2020distributed}  & Security  & Second order optimization & Defence Method & Newton’s  method \\ \cline{1-5}
				
				\cline{1-5}
				\cline{1-5}
				\bottomrule
			\end{tabular}
		\end{center}
		%}
	\caption{Privacy and security in FL.}
	\label{Tab:privacy}
\end{table*}
\normalsize

\subsection{Model Security}

Due to the privacy concern, the dataset as well as the training process of the participants is invisible to the server, which leads to considerable security risk. Specifically, malicious participants (referred to as Byzantine participants) can send elaborately designed parameters to the servers to influence the performance of the finally obtained model, e.g.,  converging to ineffective sub-optimal models or causing targeted poisoning~\cite{bhagoji2019analyzing}. In the following part of this section, we will first introduce the attack method in terms of security issues in the FL, and then we will introduce the defence method against this attack methods.

\para{Attack Method.}
Due to the different goals of the malicious participants, their attack methods can be divided into untargeted performance attack and targeted poisoning attack, which are introduced in details as follows:
\begin{packed_itemize}

\item \textbf{Untargeted performance attack:} This kind of attacks aims at reducing the overall performance of the obtained model through FL. 

\item \textbf{Targeted poisoning attack:} This kind of attacks aims at injecting a backdoor to the obtained model, which leads to its misclassification on a target sample set.
\end{packed_itemize}

%\para{Data poisoning attack:} 

Bhagoji et al.~\cite{bhagoji2019analyzing} focus on the targeted poisoning attack. Specifically, they propose two indicators of stealth to detect Byzantine participants, and their further incorporate these two indicators in the optimization object of the adversary to derive a stealthy poisoning strategy. Their experiments should that Byzantine-resilient aggregation strategies~\cite{blanchard2017machine,yin2018byzantine}, which are designed for defending the untargeted performance attack, are not robust to the proposed attacks.

\para{Defence Method.} The defence methods against the attacks of model security are mainly based on designing different aggregation rules of the model updated by the participants.

Blanchard et al.~\cite{blanchard2017machine} focus on the untargeted performance attack, and they propose the concept of $(\alpha,f)$-Byzantine resilience, where $f$ is the maximum tolerant number of Byzantine participants, and $\alpha$ characterizes the maximum tolerant distance between the optimal model and obtained model under the attack of Byzantine participants. Authors in~\cite{blanchard2017machine} further propose a aggregation rule to achieve $(\alpha,f)$-Byzantine resilience. Specifically, this aggregation rule defines a score function based on the distance between the model parameter vector updated by each participant and its $n-f-2$ closest model parameter vectors, and then the model parameter vector with the smallest score is selected to be the aggregation result.
Yin et al.~\cite{yin2018byzantine} propose two distributed gradient descent algorithms based on media and trimmed mean operations, and they futher show that these algorithms have strong ability in defense against Byzantine failures in terms of both experiments and theoretical proofs, where their statistical error rates for strongly convex, non-strongly convex, and non-convex population loss functions are established.
Li et al.~\cite{li2020byzantine} focus on federated multi-task learning against Byzantine agents, and they propose an online weight adjustment rule based on the accumulated loss to measure the similarities among agents, which is shown to be more robust to Byzantine agents than traditional distant based on similarity measurement.
Ghosh et al.~\cite{ghosh2020distributed} propose a federated optimization algorithm based on Newton's method, which is communication-efficient as well as robust against Byzantine failures.

\section{Application}\label{app}
As we have seen,  the FL  applications have witnessed the rapid development, and we can expect this trend will continue. In this section, we will introduce the applications of FL from the perspectives of both technical and real applications.

\subsection{Technical applications}
\para{Recommendation.}
Recommendation techniques have been widely
used to model user interests and mitigate information overload problems in many real applications\cite{silva2019federated,10.1145/3178876.3185994,10.1145/3570959}. For further improving privacy and security, FL has been introduced in recommendation \cite{arxiv.2301.00767}. HegedHus et al.~\cite{hegedHus2019decentralized} compare the performance of the recommendation based on FL with centralized structures and the gossip learning with decentralized structures, and show that they have a comparable performance.
Wu et al.~\cite{wu2021fedgnn} propose a privacy-preserving recommendation method based on federated GNN, where the privacy is protected based on generating pseudo interacted items and add perturbation to the transimitted embedding vectors based on local differential privacy. In addition, high-order user-item interactions are modeled by matching neighboring users based on the homomorphic encryption technique.
Qi et al.~\cite{qi2020privacy} propose a privacy-preserving new recommendation method, which protects user privacy based on clipping gradients and adding a Laplace noise to the clipped gradients before aggregation.
Liu et al.~\cite{liu2021fedct} focus on utilizing the FL techniques to solve collaborative transfer learning problem between different service providers. Specifically, they solve this problem by keeping the CF prediction models private for different domains and using an intermediate transfer agent to aggregate user information across domains.
Ammad et al.~\cite{ammad2019federated} propose a federated collaborative filtering algorithm for privacy-preserving recommendation.
Flanagan et al.~\cite{flanagan2020federated} propose a federated multi-view matrix factorization method to avoid collecting raw user data centralizedly by collaborating between the FL server, item server, and clients.
Gao et al.~\cite{gao2019privacy} design a privacy-preserving cross-domain location recommendation by proposing a Laplace perturbation mechanism using both semantics and location information.
Gao et al.~\cite{gao2020dplcf} propose a differential private local CF algorithm based on the asymmetric flipping perturbation mechanism.

\para{Natural Language Processing.} 
Natural language processing (NLP) is extensive in mobile applications. To support various language understanding tasks with privacy-preserving, a foundation NLP model is often fine-tuned in a federated setting \cite{arxiv.2212.05974}. Chen et al.~\cite{chen2021fedmatch} propose FedMatch, a federated Question Answering (QA) model, which partitions the model into two components of backbone and patch and share only the backbone between different participants to distill the common knowledge while keep the patch component private to retain the domain information for different  participants to overcome the data heterogeneity. What's more, they propose four different  patch insertion ways and two types of patch architectures.

\para{Spatio-Temporal Modeling.} In real world, particularly in climate and environmental problems, we often have measurements at a number of locations in the specific time, and will often need to make predictions of the output at unmeasured locations  in the next several time steps \cite{hasegawa2022global}.
We can approach the spatial problem by constructing spatio-temporal modeling via FL. Meng et al.~\cite{meng2021cross} propose a federated GNN-based method to predict traffic flow. Specifically, an encoder-decoder model is trained based on FedAvg to extract node-level temporal dynamics. Then, a GNN model in the server is used to model the uploaded embeddings of the node-level temporal dynamics to further extract the spatial dynamics based on split learning. Finally, the local encoder-decoder model is able to incorporate both temporal and spatial dynamics to implement the traffic flow prediction.
Feng et al.~\cite{feng2020pmf} propose a federated mobility prediction neural network model, which adds Laplace noise to the location data to protect user privacy. In addition, they propose to divide the neural network into sub-modules with different groups by their privacy leakage risk level, and train different groups of modules with different privacy protection levels.

\iffalse
\para{Communication Network Modeling.} 
Traffic flow prediction~\cite{meng2021cross}
\fi

\para{Knowledge Graph.} Knowledge graphs (KGs), consisting of  a large number of triple data with the form of (head, relation, tail),  have become essential data supports for an increasing number of applications. For further enhance decentralized privacy-preserved applications of KGs, FL has been incorporated \cite{DBLP:conf/ijcai/Chen0YCDHC22}. Chen et al.~\cite{chen2020fede} propose a FL algorithm to complete multiple knowledge graph without sharing their private triples, where they propose an optimization-based model fusion procedure at clients. Peng et al.~\cite{Peng2021Differentially} propose a differential private federated algorithm to learn the knowledge graph embeddding, which utilizes a module named privacy-preserving adversarial translation (PPAT) to project embeddings of aligned entities and relations of different knowledge graph to capture their relationship. Specifically, this PPAT module is designed based on the technique of PATE-GAN~\cite{jordon2018pate} to protect user privacy. 

\subsection{Real applications}

\para{Human Health.} As is known, the human health record data is highly private and sensitive, and thus FL can be  naturally fit for the cooperation in medical machine learning tasks across multiple institutions \cite{DBLP:journals/corr/abs-2211-07300}.  Liu et al.~\cite{liu2020experiments} compare the performance of federated algorithms with different models to detect COVID-19 based on  chest X-ray images, of which the models include MobileNet, ResNet18, and COVID-Net. Brisimi et al.~\cite{brisimi2018federated} focus on utilizing the sparse Support Vector Machine to predict hospitalization based on FL. Specifically, they propose an iterative cluster Primal Dual Splitting (cPDS) algorithm to sovle this problem. Yang et al.~\cite{yang2021FLOP} propose a FL algorithm automatic diagnosis of COVID-19 and protect patient privacy simultaneous. Specifically, they propose an algorithm named Federated Learning on medical datasets using partial networks (FLOP), which shares only a partial model between the server and clients to protect user privacy.
Silva et al.~\cite{silva2019federated} focus on investigating brain structural relationships with diseases based on FL techniques. Specifically, they propose a framework for data standardization, confounding factors correction, and multivariate analysis based on ADMM and federated principal component analysis (fPCA).

\para{Smart Transportation.} With the large-scale urbanization, it is urgently needed to intelligentize the
vehicular networks (VANET) so as to mitigate the traffic congestion, and  FL has become a promising technique due to its natural fit for the network topology \cite{DBLP:conf/meditcom/ElbirSCGB22,9360666}. Liu et al. \cite{9082655} mitigates the privacy and security issues of the existing data gathering fashion by combining FL with gated recurrent unit neural network (FedGRU) for traffic flow prediction. For further mining the spatio-temporal information for traffic flow prediction, Yuan et al. \cite{9737410} first proposes to transform the traffic trajectory  into a graph representation, and then combines the FL paradigm,  mitigating the privacy issue in centralized paradigm. Meng et al. \cite{10.1145/3447548.3467371} propose a cross-node FL combining with GNN approach,  while it  fully exploits the spatial and temporal dynamics of the server and devices, it allows the data generating on each node

\scriptsize
\begin{table*}[ht]
	%  \resizebox{1\textwidth}{!}{%
		\begin{center}
			\begin{tabular}{l|l|l}
				\toprule
				\textbf{Ref.} & \textbf{Applications} &\textbf{Idea}   \\
				\midrule
				Comparison~\cite{hegedHus2019decentralized}        & Recommendation &Comparison between FL and gossip learning.  \\ \cline{1-3} 
				CNFGNN~\cite{meng2021cross}        & Traffic flow prediction & Modeling spatial dynamics based on federated graph neural network \\ \cline{1-3} 
				FedGNN~\cite{wu2021fedgnn}        & Recommendation & Generating pseudo interacted items, FE-based neighboring users matching  \\ \cline{1-3} 
				FedNewsRec~\cite{qi2020privacy}    & Recommendation & Gradient clipping, adding Laplace noise \\\cline{1-3} 
				% &        &  &  &   & \\ \cline{2-6} 
				FedCT~\cite{liu2021fedct}         & Recommendation    &Cross-domain transfer learning, intermediate transfer agents on user devices. \\ \cline{1-3} 
				FED-MNMF~\cite{flanagan2020federated}  & Recommendation  &  Collaborating between the FL server, item server, and clients \\ \cline{1-3} 
				FCF~\cite{ammad2019federated}  &  Recommendation & Federated collaborative filtering. \\ \cline{1-3} 
				FLOP~\cite{yang2021FLOP} & Automatic diagnosis  & Sharing a partial model between the server and clients. \\ \cline{1-3} 
				FedMatch~\cite{chen2021fedmatch} & Question answer & Partition the model into the shared backbone and the private patch.  \\ \cline{1-3} 
				FedE~\cite{chen2020fede} & Knowledge graph completion & Optimization-based model fusion procedure at clients.  \\ \cline{1-3} 
				FKGE~\cite{Peng2021Differentially} & Knowledge graph embedding & PATE-GAN based translation module between different knowledge graphs. \\ \cline{1-3} 
				Experiments~\cite{liu2020experiments}   & COVID-19 detection & Compare different models. \\ \cline{1-3} 
				SegCaps~\cite{kumar2021blockchain} & COVID-19 detection & Data normalization, capsule network, blockchain. \\ \cline{1-3} 
				cPDS~\cite{brisimi2018federated}   & Hospitalization prediction & SVM, iterative cluster primal dual splitting algorithm. \\ \cline{1-3} 
				fPCA~\cite{silva2019federated} & Brain structure analysis  & ADMM, data standardization, confounding factors correction, and multivariate analysis. \\ \cline{1-3} 
				CCMF~\cite{gao2019privacy} & Recommendation & Laplace perturbation mechanism using both semantics and location information. \\ \cline{1-3} 
				DPLCF~\cite{gao2020dplcf} & Recommendation & Asymmetric flipping perturbation mechanism. \\ \cline{1-3} 
				\cline{1-3}
				\bottomrule
			\end{tabular}
		\end{center}
		%}
	\caption{ List of FL applications.}
	\label{Tab:fedapp}
\end{table*}
\normalsize
	\section{Challenges and Opportunities}\label{cao}

We have reviewed the major topics in FL. Next, we will provide the future research directions of FL, with the challenges and opportunities.  They can be summarize as follows:
\begin{packed_itemize}
    \item \textbf{FL with knowledge distillation:} The knowledge distillation technology has a strong ability to fuse a large number of deep neural network models into one single efficient model by transferring knowledge between them. A number of studies have sought to combine the knowledge distillation technology to design more efficient FL method~\cite{he2020group,lin2020ensemble}. Based on the knowledge distillation technology, we are able to develop more efficient model aggregation mechanism at the server, or design more efficient transmitted parameters between the participants and server, which are able to further fused with the privacy-preserving mechanisms to better protect the privacy of the participants.
    \item \textbf{FL for digital twin paradigm:} The rising paradigm of digital twin requires accurately simulating physical objects. However, building the simulating model requires real-world data of the target physical objects, which leads to privacy concerns. The FL techniques provide a promising solution to this problem. Specifically, we can train the simulation models for physical objects at end devices or edge devices close to the data collector of the physical objects. Then, the FL methods enable us to accurately train the model based on the massive data collected from multiple physical objects without leaking private data of each physical object, which paves the way for constructing an accurate and effective digital twin system.
    \item \textbf{High performance theoretical approach:} The massive clients normally generate the data on their own fashion and the data is generally non-IID. Moreover, in most cases, since only a small portion of clients participate in the update, this will lead to model divergence problem. SCAFFOLD and Mime have been proposed for handing this problem, both with the main idea of estimating the global gradient \cite{SCAFFOLD, MIME}. Therefore,  we can develop a  more compact estimate of the global gradient, where the global gradient estimate is aggregated on the server and also updated on  the clients, this will reach a performance improvement. 
    \item \textbf{Decentralized FL over network:} As is known, the most popular FL architecture is centralized. However, in many real world applications, the clients are decentralized massively, they frequently communicates with each other without a server coordinating them, forming a complicated network. Hence, we are in urgent need of semi-decentralized or  decentralized algorithms to cope with the increasingly complex networks. Very recently, researchers start to study decentralized FL in theoretical aspects \cite{decen1,decen3,https://doi.org/10.48550/arxiv.2203.02865}. However, decentralized FL includes the asynchronous local training, and the network dynamics make the clients participating and quitting the training frequently. These technical obstacles will bring significant challenges to the decentralized algorithm development. 
    
\end{packed_itemize}
 	\bibliographystyle{ACM-Reference-Format}
 	\bibliography{Introduction,Survey,Architecture_cgj,Model,Optimization,Privacy,Application}
		
	\end{document}